\documentclass{article}

\usepackage{fullpage}
\usepackage[utf8]{inputenc} 
\usepackage[T1]{fontenc}    
\usepackage{hyperref}       
\usepackage{url}            
\usepackage{booktabs}       
\usepackage{amsfonts}       
\usepackage{nicefrac}       
\usepackage{microtype}      
\usepackage{xcolor}         
\usepackage{amsmath, amssymb, graphicx, url}
 \usepackage{algorithm}
 \usepackage{algpseudocode}
\usepackage{times}
\usepackage{subfigure}
\usepackage{tikz,pgfplots}
\usetikzlibrary{shapes}
\usepackage{enumitem}
\usepackage{float}
\pgfplotsset{compat=newest}
\usepackage{xcolor}
\hypersetup{
	colorlinks,
	linkcolor={red!40!gray},
	citecolor={blue!40!gray},
	urlcolor={blue!70!gray}
}

\newif\ifisbold
\isboldfalse

\ifisbold\def\g {\mathbf{g}}\else\def\g{ g}\fi %

\def\Ec{{\mathcal E}}

\ifisbold \else \fi%

\def\Dc{\mathcal{D}}

\def\Fc{\mathcal{F}}
\def\Xc{\mathcal{X}}

\ifisbold\else\fi %

\def\R{\mathbf R}
\ifisbold \def\H{\mathbf H} \else \def\H{H}\fi %

\ifisbold\def\K{\mathbf K}\else\def\K{ K}\fi %

\ifx\BlackBox\undefined
\newcommand{\BlackBox}{\rule{1.5ex}{1.5ex}}  
\fi
\DeclareMathOperator*{\argmin}{\mathop{\mathrm{argmin}}}

\DeclareMathOperator*{\diag}{\mathop{\mathrm{diag}}}
\def\x{ x} 

\ifisbold\def\y{\mathbf y}\else\def\y{ y}\fi%

\ifisbold\else\fi%
\ifisbold\def\a{\mathbf a}\else\def\a{ a}\fi
\ifisbold\def\b{\mathbf b}\else\def\b{ b}\fi

\ifisbold\def\v{\mathbf v}\else\def\v{ v}\fi

\ifisbold\else\fi %

\ifisbold \else\fi %

\ifisbold \def\B{\mathbf B}\else\def\B{ B}\fi %

\ifisbold\def\A{\mathbf A}\else\def\A{ A}\fi %
\ifisbold\else\fi %
\ifisbold\def\U{\mathbf U}\else\def\U{ U}\fi %

\ifisbold\def\D{\mathbf D}\else\def\D{ D}\fi %
\ifisbold\else\fi %
\ifisbold\else\fi %

\ifisbold \def\S{\mathbf S}\else\def\S{ S} \fi%

\ifisbold\else\fi %

\ifisbold\def\I{\mathbf I}\else\def\I{ I}\fi %

\ifisbold \def\A{\mathbf A}\else\def\A{ A}\fi %

\ifisbold\def\P{\mathbf P}\else\def\P{ P}\fi
\ifisbold\def\Q{\mathbf Q}\else\def\Q{ Q}\fi %

\def\E{\mathbb E}

\def\R{\mathbb R} 
\def\tr{\mathrm{tr}}

\let\origtop\top
\renewcommand\top{{\scriptscriptstyle{\origtop}}} 

\definecolor{silver}{cmyk}{0,0,0,0.3}
\definecolor{yellow}{cmyk}{0,0,0.9,0.0}
\definecolor{reddishyellow}{cmyk}{0,0.22,1.0,0.0}
\definecolor{black}{cmyk}{0,0,0.0,1.0}
\definecolor{darkYellow}{cmyk}{0.2,0.4,1.0,0}
\definecolor{darkSilver}{cmyk}{0,0,0,0.1}
\definecolor{grey}{cmyk}{0,0,0,0.5}
\definecolor{darkgreen}{cmyk}{0.6,0,0.8,0}

\newcommand{\Green}[1]{{\color{darkgreen}  {#1}}}
\newcommand{\Blue}[1]{\color{blue}{#1}\color{black}}
\newcommand{\Brown}[1]{{\color{brown}{#1}\color{black}}}

\ifx\proof\undefined
\newenvironment{proof}{\par\noindent{\bf Proof\ }}{\hfill\BlackBox\\[2mm]}
\fi

\ifx\theorem\undefined
\newtheorem{theorem}{Theorem}
\fi

\ifx\example\undefined
\newtheorem{example}{Example}
\fi

\ifx\property\undefined
\newtheorem{property}{Property}
\fi

\ifx\lemma\undefined
\newtheorem{lemma}{Lemma}
\fi

\ifx\proposition\undefined
\newtheorem{proposition}{Proposition}
\fi

\ifx\remark\undefined
\newtheorem{remark}{Remark}
\fi

\ifx\corollary\undefined
\newtheorem{corollary}{Corollary}
\fi

\ifx\definition\undefined
\newtheorem{definition}{Definition}
\fi

\ifx\conjecture\undefined
\newtheorem{conjecture}[theorem]{Conjecture}
\fi

\ifx\axiom\undefined
\newtheorem{axiom}[theorem]{Axiom}
\fi

\ifx\claim\undefined
\newtheorem{claim}[theorem]{Claim}
\fi

\ifx\assumption\undefined
\newtheorem{assumption}{Assumption}
\fi

\newcommand {\michael}[1]{{\ifishidden\else\color{red}\sf{[Michael: #1]}\fi}}

\newif\ifishidden
\ishiddentrue

\newcommand{\hide}[1]{{\ifishidden\else\textcolor{orange}{[HIDE] #1}\fi}}

\begin{document}

\title{Recent and Upcoming Developments in \\ Randomized Numerical Linear Algebra for Machine Learning}

\author{Micha{\l} Derezi\'nski\thanks{University of Michigan
    (\texttt{derezin@umich.edu})}
  \and Michael W. Mahoney\thanks{ICSI, LBNL, and University of
    California, Berkeley (\texttt{mmahoney@stat.berkeley.edu})}}

\maketitle
\begin{abstract}
\noindent
Large matrices arise in many machine learning and data analysis applications, including as representations of datasets, graphs, model weights, and first and second-order derivatives. 
Randomized Numerical Linear Algebra (RandNLA) is an area which uses randomness to develop improved algorithms for ubiquitous matrix problems. 
The area has reached a certain level of maturity; but recent hardware trends, efforts to incorporate RandNLA algorithms into core numerical libraries, and advances in machine learning, statistics, and random matrix theory, have lead to new theoretical and practical challenges. 
This article provides a self-contained overview of RandNLA, in light of these developments.
\end{abstract}

\section{Introduction}

Matrices provide a natural structure with which to model data.
For example, a matrix $A\in\mathbb{R}^{m \times n}$ can encode information about $m$ objects, each of which is described by $n$ features.
Alternatively, a positive definite matrix $A\in\mathbb{R}^{n \times n}$ can encode correlations/similarities between all pairs of $n$ objects. 
Motivated by large-scale data problems, recent years have witnessed many exciting developments in the theory and practice of matrix algorithms.
Particularly remarkable is the use of randomization.
Historically, in statistics, machine learning (ML), and domain sciences, randomization has been assumed to be a property of the input data, e.g., due to noise in the data generation mechanisms.
In this more recent work on randomization, it is used as an algorithmic or computational~resource. 

\emph{Randomized Numerical Linear Algebra (RandNLA)} is an interdisciplinary research area that exploits randomization as a computational resource to develop improved algorithms for large-scale linear algebra problems.
From a \emph{foundational perspective}, it has roots in theoretical computer science (TCS), deep connections with convex analysis, probability theory, and metric embedding theory, etc., as well as strong connections with scientific computing, signal processing, and numerical linear algebra (NLA).
From an \emph{implementational perspective}, well-engineered RandNLA algorithms beat highly-optimized software libraries for ubiquitous problems such as very over-determined least-squares, they scale well to parallel/distributed environments, and they beat state-of-the-art for a wide range of low-rank matrix approximation problems.
From a \emph{data analysis perspective}, RandNLA has strong connections with ML and statistics and many ``non-methodological'' applications of data analysis. 
More generally, of course, it is of continued importance since there is a growing interest in providing an \textit{algorithmic and statistical foundation for modern large-scale data analysis}.

The area of RandNLA has achieved a certain level of maturity.
As such, there are multiple reviews of the area from multiple different perspectives:
introductory overviews (light on prerequisites)~\cite{DM16_CACM,RandNLA_PCMIchapter_chapter}; 
broad and proof-heavy resources~\cite{woodruff2014sketching,TroppMatrixConcIneqBOOK2015,Mah16_RLA_TR}; 
perspectives on interdisciplinary theory (light on proofs)~\cite{Mah-mat-rev_BOOK,dpps-in-randnla}; 
deep investigations of specific disciplinary topics \cite{tropp2011structure,KV17_acta,Martinsson_PCMIchapter_chapter,martinsson2020randomized}; and
approaches to high-quality software implementations~\cite{randlapack_book_v2_arxiv}.
Particularly notable is the current effort of incorporating RandNLA algorithms into the core numerical libraries (e.g., RandLAPACK and RandBLAS; see \cite{randlapack_book_v2_arxiv}) that lie at the foundation of virtually all computational tools in ML (and scientific computing and beyond). 

This level of maturity, as well as recent demands by the ML community and recent trends in hardware, lead to new theoretical and practical challenges that did not exist a decade ago. 
For example:
developing RandLAPACK and RandBLAS leads to new algorithmic and theoretical abstractions, different than those present in TCS or NLA, and different than those common in statistics and ML; 
recent developments in neural network training highlight important trade-offs between communication and computation, between forward accuracy and parameter stability, etc.; and
recent developments in hardware have led to new trade-offs between latency and throughput, both at the model training and model inference stage.
To complement these challenges, recent theoretical developments in Random Matrix Theory (RMT) have provided finer quantitative performance analysis than was possible with traditional RandNLA analysis tools.
These theoretical developments come from RMT as well as from RandNLA itself, and they permit both finer analysis of existing methods as well as the possibility to develop novel methods that better bridge the theory-practice~gap.

In this survey, we will provide a self-contained review/overview of RandNLA, in light of these recent developments, describing and highlighting upcoming and future trends.
We will introduce the foundations of RandNLA and matrix sketching, with a particular focus on applications in ML and stochastic optimization, followed by an overview of recent developments in using sketching methods to gain stronger control on the convergence and generalization properties of stochastic algorithms. 
We will cover both the theoretical aspects of these techniques, as well as their applications in the context of important ML and data science research topics. 
Thus, our discussion will be relevant not only to theoretical researchers who wish to learn the latest advances in RandNLA, but also to a general audience of ML and data science researchers and practitioners, who want to incorporate RandNLA methods into their large-scale data problems.

\section{Foundations of ``Classical'' RandNLA}
\label{sxn:classical-randnla}

In this section, we will describe the foundations of RandNLA theory, at least up until a few years ago.
The basic idea is to construct a sketch (either data-aware or more commonly data-oblivious) that has parameters chosen basically to preserve the geometry of an entire low-dimensional subspace.
This sketch can be interpreted in one of several complementary ways; and it can be used in one of several complementary ways.
Understanding the details of these complementary approaches is crucial for understanding recent advances and upcoming developments in RandNLA for modern~ML.

\subsection{Matrix Multiplication}
\label{sxn:mat-mult}

A core primitive in RandNLA is that of approximating the product of two matrices \cite{drineas2006multiplication}.
The basic problem is the following:
given an $m \times n$ matrix $A$ and an $n \times q$ matrix $B$, approximate the product $A \cdot B$.
A key conceptual insight is that this problem can 
be expressed as approximating the sum of $n$ rank-one matrices:
$$
A\cdot B = \sum_{k=1}^n \left(\begin{array}{c}
 \\
 \hspace{-1mm}
\hspace{-1mm} A_{*k} \hspace{-2mm}\\
 \hspace{-3mm}
 \\
\end{array}\right) \cdot \left(\begin{array}{cccc}
   & \hspace{-1mm} B_{k*} & \hspace{-2mm} \\
\end{array}\right)  .  
$$

\noindent
Given this, 
a natural (random sampling) algorithm suggests itself:
\begin{algorithmic}[1]
\State
Fix a set of probabilities $p_i$, $i=1,\ldots,n$, summing up to $1$.\vspace{1mm}
\For{$t=1,\ldots,s$}\vspace{1mm}
\State
Set $j_t = i$ according to $\mathbb{P}[j_t = i] = p_i$.\vspace{1mm}

\EndFor \quad (Randomly pick $s$ terms of the sum according to the $p_i$'s.) \vspace{1mm}
 \State
Approximate 
$AB$ by summing the $s$ terms, after scaling.\vspace{1mm}
\end{algorithmic}

\hide{Here is the same algorithm, but in matrix notation:
\begin{enumerate}
\item
Pick $c$ columns of $A$ to form an $m \times c$ matrix $C$ and the corresponding $c$ rows of $B$ to form a $c \times p$ matrix $R$.
\item
Rescale the columns/rows prior to including them in $C$/$R$.
\item
Approximate $A B$ by $C R$.
\end{enumerate}
}

\noindent
If we let $S$ be an $s \times n$ matrix whose $t^{\mathrm{th}}$ row
($t=1,...,s$) has one non-zero, 
$
S_{t, j_t} = 1/\sqrt{sp_{j_t}}
$,
then we can use a ``sampling matrix'' formalism
\cite{drineas2006multiplication} (a type of ``sketching
operator'') to express this as:
$$
AB \approx
\frac{1}{s}\sum_{t=1}^{s}\frac{1}{p_{j_t}}a_{j_t}b_{j_t}^\top =
\left(AS^\top \right) \cdot ( S B) =  C\cdot R  ,
$$
where 
$a_{j_t}$ and $b_{j_t}^\top$ are corresponding columns of $A$ and rows of~$B$.
Now, if we use probabilities $p_i \propto
\|a_i\|_2\|b_i\|_2$ (row-norm sampling
\cite{drineas2006multiplication}), then we can minimize a very natural
Frobenius norm error:
\begin{equation}
\mathbb{E}\left[ \|AB-CR\|_F \right] 
   = \mathbb{E}\left[ \|AB-(AS^\top)(SB)\|_F \right]
   \le \frac{1}{\sqrt{s}} \|A\|_F \|B\|_F  .
\label{eqn:basic_mm_result}
\end{equation}

\noindent
Here, by setting $s = 1/\epsilon^2$, we obtain a Frobenius norm error that is bounded above by $\epsilon$, in the additive scale defined by $\|A\|_F \|B\|_F$.

This \textit{Frobenius norm} bound given in (\ref{eqn:basic_mm_result}) is used in \textit{many} places in RandNLA.
However, ``better'' \textit{spectral norm} bounds of the form
$\|AB-CR\|_2\leq\epsilon \|\A\|_2\|\B\|_2$  are possible (after
adjusting the sample size $s$; see \cite{cohen2016optimal}) via
matrix versions of Chernoff/Bernstein concentration inequalities
\cite{TroppMatrixConcIneqBOOK2015}.

Within RandNLA, the main ``use case'' for approximate matrix
multiplication arises when it is applied not directly to the matrices themselves, but rather to the ``subspaces'' that they
define, giving rise to \emph{subspace embeddings} \cite{Mah-mat-rev_BOOK,woodruff2014sketching}.

\begin{definition}\label{def:subspace_embedding}
Let $\A$ be an $n\times d$ matrix, and let
$\U$ consist of orthonormal columns that are the
basis of the column subspace of $\A$. 
Then, an $s\times n$ matrix $S$ is a \textit{subspace embedding} for
$\A$ if
\begin{equation*}
\| U^\top U-\left(S U\right)^\top S U \|_2
   = \| I-\left(S U\right)^\top S U \|_2 \le \epsilon .
\label{eqn:subspace_embed_def}
\end{equation*}
\end{definition}
\noindent
Here, for a sampling matrix $\S$, the ``correct'' sampling
probabilities $p_i$ to use are proportional to 
the squared row norms $\|U_{i*}\|_2^2$, a.k.a. the \emph{leverage scores} of
$\A$.
These are expensive to compute exactly, but they are fast to approximate---explicitly or
implicitly---in one of many ways
\cite{fast-leverage-scores,ailon2009fast,cw-sparse,nn-sparse,mm-sparse}.

Subspace embeddings are an extremely important concept in Classical RandNLA theory.
They were introduced and first used implicitly in RandNLA by
\cite{drineas2006sampling,cur-decomposition}, who relied on leverage
score sampling \cite{fast-leverage-scores}, which can be viewed as a
type of \emph{data-aware sketching}.
The first explicit definition of subspace embeddings was given by \cite{sarlos-sketching,DMMS07_FastL2_NM10}, who focused on \textit{data-oblivious} sketching; and
these data-oblivious methods were popularized by \cite{woodruff2014sketching}.
Many constructions (random sampling and projection methods, deterministic constructions, hashing functions, etc.) satisfy this condition \cite{woodruff2014sketching}.

There are several complementary interpretations of a subspace embedding.
Within NLA, it is an \emph{acute perturbation} \cite{Ste77}, meaning in particular that distances and angles are perturbed, but rank is not lost.
Within TCS, it is a Euclidean space analogue of \emph{Johnson-Lindenstrauss (JL) lemma} \cite{Matousek08_RSA}, meaning in particular that distances are approximately preserved for the infinite number of points on a unit ball in the low-dimensional space.
More generally, the subspace embedding guarantee can be used to
produce a \emph{spectral approximation} of the matrix product
$\A^\top\A$, in terms of the Loewner ordering of positive
semi-definite (PSD) matrices, which provides a strong control on \emph{all}
the associated eigenvectors and eigenvalues. 
To be more precise,
  let 
  $\A$ be an $m\times n$ matrix and $\epsilon\in[0,1/2]$. If
  matrix $\S$ is an $\epsilon/2$ subspace embedding for the
  subspace defined by $\A$; then
$\A^\top\S^\top\S\A$ is a spectral approximation of $\A^\top\A$, i.e.,
\begin{align}
\frac1{1+\epsilon}\A^\top\A \preceq \A^\top\S^\top\S\A \preceq (1+\epsilon) \A^\top\A,
\label{eq:spectral}
\end{align}
where $\preceq$ denotes the PSD Loewner ordering. 
(We use $\A^\top\S^\top\S\A\approx_{\epsilon}\A^\top\A$ as a shorthand for \eqref{eq:spectral} throughout.)
The property 
$\A^\top\S^\top\S\A\approx_{\epsilon}\A^\top\A$ implies
that any $i^{\mathrm{th}}$ eigenvalue of the sketch $\A^\top\S^\top\S\A$ is an $\epsilon$
approximation of the corresponding eigenvalue of the data matrix
$\A^\top\A$ (and similarly for the singular values of $\S\A$). It also
extends to the control of various matrix functions of $\A^\top\A$,
including the inverse, i.e.,
$(\A^\top\S^\top\S\A)^{-1}\approx_{\epsilon}(\A^\top\A)^{-1}$, square
root, trace, quadratic form, and more.

This subspace embedding property is a ``must must have'' for the worst-case style of analysis provided by TCS.
For everyone else, it provides a guiding principle, but (strictly speaking) it's optional.
For example, 
if one is interested in providing good numerical implementations, then losing rank can still give a good preconditioner \cite{blendenpik,ANT09}; and
if one is interested in statistics and ML, then losing rank introduces a bit of bias that can be compensated for by greatly reducing variance \cite{MMY15,GarveshMahoney_JMLR,ma2022asymptotic}.
\emph{This leads to a theory-practice gap.
As we will describe in Sec.~\ref{sxn:modern-randnla}, this is the key technical place where Classical RandNLA theory and Modern RandNLA theory differ, and why recent developments in Modern RandNLA theory open up the door for so many other use cases and applications of RandNLA ideas.
}

\subsection{Least-squares Approximation}
\label{sxn:least-squares}

We can use ideas from Sec.~\ref{sxn:mat-mult} for RandNLA methods for approximate matrix multiplication as a foundation to develop RandNLA methods for least-squares (LS) regression.
Given $m \times n$ matrix $A$ and $m$-dimensional vector $b$, the basic LS problem is to solve 
\begin{equation*}
\label{eqn:LSmotivation}
x^* = \argmin_{x \in \mathbb{R}^n} \| Ax - b \|_2  .
\end{equation*}

\noindent
If $m \gg n$, then the problem is called overdetermined or overconstrained; and 
if $m \ll n$, then the problem is called underdetermined or underconstrained (or, especially in ML, overparameterized).
In the overdetermined case, one can compute the solution in $O(mn^2)$ time
(in RAM model), e.g., with normal equations, QR
decompositions, or the Singular Value Decomposition (SVD). 
RandNLA gives faster algorithms for this ubiquitous problem, which is at the
root of most other advances (in particular, work on low-rank matrix approximation, as well as extensions to optimization problems, etc.) in the area. 

Importantly, depending on the domain, ``faster'' can mean one of several different things:
within TCS, 
``faster'' is in terms of low-precision asymptotic worst-case theory; 
within NLA, 
``faster'' is in terms of high-precision wall-clock time;
and
in large-scale implementations, 
one can compute (in Spark/MPI/etc.) low, medium, and high precision solutions on up to terabyte-sized data, and here ``faster'' depends mostly on communication.

\vspace{1mm}\noindent\textbf{Meta-algorithm for LS regression.}
Here is a meta-algorithm for solving LS regression problems with RandNLA methods.
\begin{algorithmic}[1]
  \State Randomly sample a small number of constraints according to
       $\A$'s leverage score sampling probabilities $\{p_i\}_{i=1}^{m}$.\vspace{1mm}
   \State Solve the LS regression problem on the resulting subproblem.\vspace{1mm}
\end{algorithmic}

\noindent
A \emph{\textit{na\"{i}ve} version} of this meta-algorithm is \emph{not promising}: 
it gives a $1+\epsilon$ relative-error approximation (since it approximates the solution using only a subset of constraints),
that fails with probability $\delta$ (comparable to flipping a fair coin
``heads,'' say, $30$ times in a row),
in roughly $O(mn^2)$ time, i.e., as long as the exact deterministic method (since, as stated, it computes the leverage scores exactly).

On the other hand, a \emph{\textit{non-na\"{i}ve} version} of this meta-algorithm is \emph{very promising}: 
it gives the best worst-case algorithm in RAM \cite{sarlos-sketching,DMMS07_FastL2_NM10,fast-leverage-scores} (using Sketch-and-Solve, described below);
it beats LAPACK for high precision in wall-clock time \cite{rokhlin2008fast,blendenpik,MSM14_SISC} (using Sketch-and-Precondition, described below);
it leads to super-terabyte-scale implementations in parallel/distributed environments \cite{YMM16_PIEEE,Git16A};
and
it gives the foundation for low-rank approximations and the rest of RandNLA \cite{drineas2006sampling,cur-decomposition,Mah-mat-rev_BOOK,RandNLA_PCMIchapter_chapter}.

\vspace{1mm}\noindent\textbf{Fundamental structural result.}
For LS approximation, here is the fundamental structural result that is at the heart of RandNLA-based methods \cite{DMMS07_FastL2_NM10,Mah-mat-rev_BOOK}.
(This is a deterministic result that holds for any matrix $S$, and it is central to RandNLA theory by choosing $\S$ to be an appropriate sketching operator.)
\hide{
Consider the over-determined LS approximation problem:
$$
\mathcal{Z}_2^2 = \min_{x\in\mathbb{R}^{n}} \| b - Ax \|_2^2 = \| b - Ax^* \|_2^2
$$
}
Consider the ``sketched'' LS approximation problem: 
$$
\tilde\x = \argmin_{x\in\mathbb{R}^{n}} \| S(\A\x-\b) \|_2 ,
$$
where $S$ is \textit{any} matrix.
If $S$ satisfies the two basic conditions:
\begin{eqnarray*}
  \text{1)}\ \  \sigma_{\min}^{2} (\S U_A ) \ge 1/\sqrt{2}
  \quad\text{and}\quad\text{2)} \ \ 
\| U_A^\top \S^\top \S b^{\perp} \|_2 \le \sqrt{\epsilon/2}\, \|\A\x^*-\b\|_2 ,
\end{eqnarray*}
where $b^{\perp} = b - U_AU_A^\top A$ and $\U_A$ is the orthonormal
basis of $\A$,  then:
\begin{eqnarray*}
\| A \tilde{x} -b \|_2 &\le& (1+\epsilon) \|\A\x^*-\b\|_2   .       
\end{eqnarray*}

\noindent
Importantly, each of the two basic conditions can be viewed
as approximate matrix-matrix multiplication
\cite{Mah-mat-rev_BOOK,RandNLA_PCMIchapter_chapter}.

RandNLA provides a plethora of sketching methods (random sampling and projection methods, deterministic constructions, hashing functions, etc.) for $\S$ so as to satisfy these two basic conditions.
As described in the many reviews of RandNLA, 
the choice of which method one uses typically depends on one's goals: 
to provide TCS-style worst-case theory;
to provide NLA-style implementations;
to use in large-scale and/or ML settings;
etc.

\begin{figure}[t] 
\centering\includegraphics[width=0.6\textwidth]{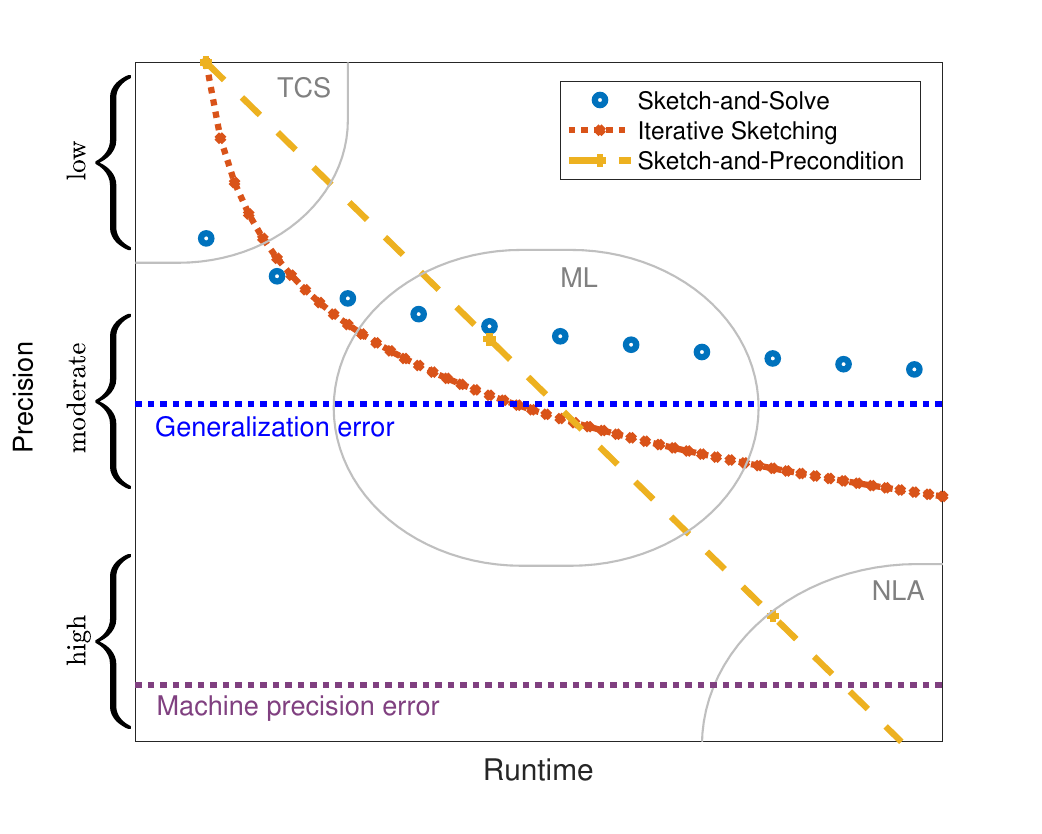}
\vspace{-2mm}
  \caption{Schematic of three algorithmic paradigms for the use of RandNLA methods:
``Classical'' (subspace embedding based) RandNLA theory is most appropriate for low precision (with Sketch-and-Solve) and high precision (with Sketch-and-Precondition); while
``Modern'' (RMT-based) RandNLA theory is well-suited for moderate precision (with Iterative Sketching), which is of increasing interest in modern ML applications.
}
\label{fig:schematic-three-paradigms}
\end{figure}

\vspace{1mm}\noindent\textbf{Three algorithmic paradigms for LS and other RandNLA methods.}
If we want to use RandNLA methods more generally, then there are three general paradigms---that apply both to LS methods as well as more broadly than LS (e.g., to low-rank matrix approximation).
\begin{enumerate}
  \item 
  \emph{Sketch-and-Solve}: 
  Here \cite{sarlos-sketching,DMMS07_FastL2_NM10,fast-leverage-scores}, we randomly construct a \textit{smaller} LS problem, and then solve it using a traditional NLA method.
  This is the simplest approach to highlight the structure of RandNLA theory; 
  it is most convenient for TCS-style theory; and 
  it leads to low-precision estimates, e.g., $\epsilon = 0.1$.
  \item 
  \emph{Iterative Sketching}: 
  Here \cite{gower2015randomized,PW16_IHS,yang2017weighted}, we repeatedly sketch/sub-sample the problem randomly, and then iteratively refine the estimate using a convex optimization method.
  This includes Preconditioned Weighted SGD, Sketch-and-Project, and
  Newton Sketch; 
  it is the most convenient for ML-style stochastic optimization; and 
  it leads to moderate precision estimates, e.g., $\epsilon = 10^{-3}$.
  \item 
  \emph{Sketch-and-Precondition}: 
  Here \cite{rokhlin2008fast,blendenpik,MSM14_SISC}, we randomly construct an \textit{equivalent} but well-conditioned problem, and then solve it using a traditional deterministic NLA iterative method.
  This is the best (usually) for high-quality numerical solutions; 
  it is most convenient for NLA-style implementations; and 
  it leads to high-precision solutions, e.g., $\epsilon = 10^{-10}$.
\end{enumerate}
See Fig.~\ref{fig:schematic-three-paradigms} for an illustration of these three algorithmic paradigms.
The first and the third have received the greatest amount of attention within RandNLA, in part due to the (TCS and NLA based) style of theory used in Classical RandNLA.
The new theoretical developments in Modern RandNLA (described in Sec.~\ref{sxn:modern-randnla}) are particularly well-suited to the second (and for ML based theory).

\subsection{Low-rank approximation}
\label{sxn:low-rank}

We can use the ideas from Sec.~\ref{sxn:least-squares} to develop RandNLA methods for low-rank (LR) approximation.
Many details change, e.g., for LS problems the matrix is usually tall, while for LR problems both dimensions may be large, in which case there is a rank parameter.
Here, we will highlight the main high-level ideas.

\vspace{1mm}\noindent\textbf{LR matrix approximation via the SVD.}
Let $A$ denote an $m\times n$ data matrix (for simplicity, suppose
that $m\geq n$), 
and 
let
$\Q\in\R^{m\times k}$ denote an ``embedding'' between the ``high'' and ``low'' dimensions; 
then
$\hat A = \Q^\top A $ is a low-dimensional representation of the data, 
and
$\tilde A =\Q\Q^\top\A $ is a LR matrix approximation.
\michael{@michal: we are a bit inconsistent with the letters we sue for dimensions; let's discuss the best way to handle this.}
If $A$ has rank higher than $k$, then the
truncated SVD give a LR approximation, 
$$ A \approx A_k = U_k\Sigma_k V_k^\top , $$ 
composed of the top $k$ singular value-vector pairs,
where recall that we order the singular values so that $\sigma_1\geq \sigma_2\geq ...\geq\sigma_n$.
Computing this SVD costs $O(mn^2)$ time, for $m\geq n$.
As for the error of the truncated SVD, we are typically interested in how much ``information'' is lost in a truncated SVD, i.e., in considering
    \begin{align*}
      A - A_k= \sum_{i>k}\sigma_i u_i v_i^\top .
    \end{align*}
  For the spectral norm, this is:
    $ 
      \|A-A_k\|_2 = \sigma_{k+1} 
    $; 
   and 
   for the Frobenius norm, this is:
    $ 
      \|A-A_k\|_F^2 = \sum_{i>k}\sigma_i^2 .
    $ 
  For any unitarily-invariant matrix norm, $\|\cdot\|_{\xi}$, the matrix $A_k$ provides an optimal rank-$k$ approximation, in the sense that
    \begin{align*}
      A_k = \mbox{argmin}_{B\,:\,\mbox{rank}(B)=k}\|A-B\|_{\xi} .
    \end{align*}
One can also view the truncated SVD as a projection, in which case it exhibits complementary optimality properties.
In particular, if we let $ P= U_k U_k^\top$ be the projection onto the
span of $ u_1,..., u_k$, then we can use that to derive $A_k$ as a projection of $A$:
      \begin{align*}
        P A =\U_k\U_k^\top\A= \sum_{i=1}^n \sigma_i \U_k\U_k^\top u_i v_i^\top  = \sum_{i=1}^k\sigma_i u_i v_i^\top= U_k\Sigma_k V_k^\top=A_k .
      \end{align*}
In this context, the total squared projection error (using $P$ as the projection) over the columns
$\a_1,...,\a_n$ of $\A$ corresponds to the squared Frobenius norm error of the
LR approximation:
      \begin{align*}
        \sum_{i=1}^n \| a_i-P a_i\|_2^2 =    \|A-A_k\|_F^2.
      \end{align*}
These complementary perspectives have been useful in developing RandNLA LR matrix approximation methods.

\vspace{1mm}\noindent\textbf{Column subset selection.}
There is another problem, the \emph{Column subset selection problem} (CSSP), that is superficially different than LR matrix approximation via the SVD, but that has a similar underlying linear algebraic structure.
The problem is to choose a small number of \emph{actual} columns from an input matrix $A$ \cite{cur-decomposition,BMD09_CSSP_SODA,near-optimal-columns}.
For most objectives of interest, it is intractable to compute the optimal set of columns.
One version of the CSSP is the following:
instead of projecting onto singular vectors, one can select a subset
of data points, $\{a_i:i\in S\}$ for $S\subseteq \{1,...,n\}$, and
project onto their span (with slight abuse of notation, here $S$ is a
subset, while later it is a sketching operator).
If we let $P_{S}= \Q \Q^\top$ be a projection onto  the span of  $\{a_i:i\in S\}$, where $\Q$ is the embedding matrix, then the problem is the following:
      \begin{align*}
        \mbox{Find}\quad\mbox{argmin}_{S:|S|=k}\| A- P_S A \|_{F} .
      \end{align*}
Even though the CSSP is different than LR matrix approximation via the SVD, and solving this problem does not in general require computing SVD, the truncated SVD is often used as a baseline quality measure for the CSSP.

\vspace{1mm}\noindent\textbf{Fundamental Structural Result for LR Approximation.}
For LR matrix approximation, including both via the SVD and via the CSSP, here is the fundamental structural result that is at the heart of RandNLA-based methods \cite{MD16_chapter}.
(As with the fundamental structural result for LS, this is a deterministic result that holds for any matrix $\S$, and it is central to RandNLA theory by choosing $\S$ to be an appropriate sketching operator.)
Given $A \in \mathbb{R}^{m \times n}$, let $V_k \in \mathbb{R}^{n \times k}$ be the matrix of the top $k$ right singular vectors of $A$.
Let $\S  \in \mathbb{R}^{n \times s}$ ($s \geq k$) be any matrix
s.t. $V_k^\top\S $  has full rank. 
Then, for any unitarily invariant norm $\xi$, 
\begin{equation}
\|A - P_{A\S }A\|_{\xi} \leq \|A - A_k\|_{\xi} +
\|\Sigma_{k,\perp}\left(V_{k,\perp}^\top\S \right)\left(V_k^\top\S
\right)^+\|_{\xi}. 
\label{eqn:LR_structural_result}
\end{equation}
This result was introduced and first used for the CSSP
\cite{BMD09_CSSP_SODA}; and subsequently it has been used for 
improved random projections in LR matrix approximations \cite{RST09,tropp2011structure},
improved Nystr\"{o}m-based LR PSD matrix approximations \cite{revisiting-nystrom},
improved feature selection methods \cite{BMD08_CSSP_KDD,BMD09_kmeans_NIPS}, and  
improved power iteration and Lanczos based methods~\cite{MD16_chapter}.

\vspace{1mm}\noindent\textbf{Basic randomized LR approximation algorithm.}
Let's go into more detail on how this result is used for random projection algorithms for LR matrix approximation \cite{tropp2011structure}.
Given an $ m \times n$ matrix $A$, a target rank $k$, and an over-sampling parameter $p$, 
the following algorithm outputs rank-$(k+p)$ factors $U$, $\Sigma$,
and $V$, s.t. $A \approx U \Sigma V^\top$.
\vspace{3mm}
\begin{algorithmic}[1]
  \State Draw an $n \times (k+p)$ Gaussian random matrix $\S$.\vspace{1mm}
\State
Form the $n \times (k+p)$ sample matrix $Y=A \S$. \vspace{1mm}
\State
Compute an orthonormal matrix $Q$ s.t. $Y=QQ^\top Y$. \vspace{1mm}
\State
Form the small matrix $B=Q^\top A$. \vspace{1mm}
\State
Factor the small matrix $B=\hat{U}\Sigma V^\top$. \vspace{1mm}
\State
Form $U = Q\hat{U}$. \vspace{1mm}
\end{algorithmic}
The use of the Gaussian random matrix $\S$ is important here.
It is slower in RAM than other fast sketching methods, and in practice
generating Gaussian random bits can be the computational bottleneck. 
However, in scientific computing, one often works with structured
operators which can be applied quickly to an arbitrary vector 
\cite{tropp2011structure}. 
In this case, one can prove bounds of the form:
\begin{eqnarray*}
\|A-QQ^\top A\|_F &\le& \left(1+\frac{k}{p-1}\right)^{1/2}\|\A-\A_k\|_F, \\
\|A-QQ^\top A\|_2 &\le& \left(1+\sqrt{\frac{k}{p-1}}\right)\|A-\A_k\|_2 + \frac{e\sqrt{k+p}}{p}  \|\A-\A_k\|_F.
\end{eqnarray*}
These expressions quantify the idea that, w.r.t. the spectral/Frobenius norm error, the computed approximation is as good as the best LR matrix approximation (provided by the truncated SVD), plus a very little bit of additional~error.

\vspace{1mm}\noindent\textbf{Extension of basic randomized LR approximation algorithm.}
These results can be improved in various ways, and the analysis typically uses some variant of the fundamental structural result in
Eqn.~(\ref{eqn:LR_structural_result}).
Consider, for example, the block power iteration scheme:\vspace{1mm}
\begin{algorithmic}[1]
\State Draw an $n\times k$
      Gaussian random matrix $\S$. \vspace{1mm}
\State Compute the matrix $\tilde U_q = \texttt{orth}((\A\A^\top)^q\A\S)$
      via alternating application of $\A$, $\A^\top$ and orthonormalization.\vspace{1mm}
\end{algorithmic}
The basic idea is that $\Tilde U_q$ converges to the
span$(U_k)$, and $\tilde U_q \tilde U_q^\top A $ gives a good $k$-rank approx of $ A$.  
For more detail, see, e.g., Lemma 4.15 from \cite{woodruff2014sketching}. 
    For any orthonormal matrix $ \Q \in \mathbb{R}^{m \times k}$,
    a basic structural result is that
    $$
    \| A-  \Q \Q^\top A\|_2 \leq \|( A A^\top)^q\A - \Q \Q^\top ( AA^\top)^q\A \|_2^{\frac{1}{2q+1}} .
    $$
    From this, by choosing $q = \mathcal{O}(\log(n)/\epsilon)$,
    $\Q=\tilde\U_q$, and a Gaussian $\S$, we can show that with
    high probability
    $$ 
    \| A-  \Tilde U_q\Tilde U_q^\top\A\|_2 \leq (1+\epsilon) \cdot\|\A-\A_k\|_2.
    $$ 
The runtime of this method is $\mathcal{O}(mnkq + nk^2q )$, where
$mnkq$ corresponds to $O(q)$ matrix multiplications
and
$nk^2q$ is $O(q)$ orthonormalizations (in theory, one orthonormalization
is enough, but performing orthonormalization after each multiplication helps numerical
stability \cite{tropp2011structure}).
This is of interest since it leads to \textit{high precision} LR approximations in the spectral norm \cite{tropp2011structure,woodruff2014sketching,musco2015randomized,DIKM19_SIMAX}.

\section{Foundations of ``Modern'' RandNLA}
\label{sxn:modern-randnla}

In this section, we will describe relatively-recent work that provides a foundation for current and upcoming developments in RandNLA.
One surprising aspect of Classical RandNLA theory, as outlined in Sec.~\ref{sxn:classical-randnla}, is that there is actually very little use of randomness.
Essentially, all the randomness is ``filtered through'' the subspace embedding.
That is sufficient for many TCS and NLA style objectives (although with Sketch-and-Solve and Sketch-and-Precondition, the two areas use that embedding differently), but 
it is often ``overkill'' when RandNLA methods are used in broader (statistical, ML, optimization) pipelines (that typically use some form of Iterative Sketching). 
This has applications, in particular, 
in stochastic optimization and traditional statistical resampling methods, where the JL-style guarantees that come with a subspace embedding are sufficient but not necessary. 
Modern RandNLA theory basically asks for a stronger form of
``Algorithmic Gaussianization'' than the JL Lemma provides, in the
sense that sketches look more Gaussian-like than what is provided with pair-wise JL-like guarantees.
To accomplish this, we must make connections with ideas from non-asymptotic RMT, which have proven fruitful across data science and ML.

\subsection{Algorithmic Gaussianization via 
RMT 
and LESS Embeddings
}
\label{sxn:modern-randnla-RMT}

To understand the basic issue, let us revisit the subspace embedding,
where for an $n\times d$ orthogonal matrix $\U$ (a column basis matrix of
$\A$) we seek a 
matrix $\S\in\R^{s\times n}$ such that $(\S\U)^\top\S\U
\approx_\epsilon \U^\top\U=\I$. Here, $s$ must be larger than $d$; but
how much larger does it need to be?

\vspace{1mm}\noindent\textbf{The proportional regime.}
Let us consider the extreme case (which is in fact the most common) where $s \approx d$ to within a
constant factor, and suppose that $\S$ is a scaled i.i.d. Gaussian sketching matrix.
Thanks to the rotation invariance of Gaussian distribution, $S U$ is
also Gaussian, and so we can obtain (up to lower order terms): 
\begin{align}
    \sigma_{\min}( S U) \approx 1 - \sqrt{\frac{d}{s}},\qquad
    \sigma_{\max}( S U) \approx 1 + \sqrt{\frac{d}{s}}  .\label{eq:mp-law}
\end{align}
This follows from the Marchenko-Pastur (MP) law \cite{BS06}.
Such a Gaussian sketch not only gives
us sharp control on subspace embeddings, but also  on most
Classical RandNLA tasks, e.g.:
  \begin{itemize}
    \item 
    Sketch-and-Solve: If we solve $\tilde{x} = \mbox{argmin}_{x}\| S( A x- b)\|_2^2$, then
      $ 
        \mathbb{E} \| A(\tilde{ x} -  x^*)\|_2^2 = \frac{d}{s-d-1} \| A
           x^* -  b\|_2^2$ for $ s\geq d+2. 
      $ 
    \item 
    Sketch-and-Precondition: If we construct $R^{-1}$ from the QR of  $ S A$, then
      $ 
        \text{cond}( A R^{-1})\leq 6$, with high probability for $ s\geq 2d . 
      $ 
    \item 
    LR approximation: If we compute $ Q = \texttt{orth}( A S)$, then
      $ 
        \mathbb{E}\| A -  Q Q^\top A\|_F^2\leq \big(1+\frac{k}{s-k-1}\big)\cdot \| A- A_k\|_F^2$ for $ s\geq k+2.
      $ 
  \end{itemize}
From these results, traditional RandNLA-style guarantees (summarized in Sec.~\ref{sxn:classical-randnla}) are straightforward to derive.
These results are all relatively easy to show for i.i.d. Gaussian matrices; and results of this form are common in NLA and scientific computing applications of RandNLA~\cite{tropp2011structure}.
The basic question is whether we can obtain similar results with (faster) non-Gaussian sketches.

\vspace{1mm}\noindent\textbf{Inversion bias.}
To answer this question, we need a notion of sketch quality (akin to subspace embedding) that can capture the ``Gaussian-like'' behavior---\emph{in the proportional regime}. 
Working in the proportional regime, where the aspect ratio of the matrix is constant, is key here---it is much more realistic, and if the aspect ratio is not constant then JL-like results are straightforward to derive.
One way to capture this Gaussian-like behavior is through \emph{inversion bias} \cite{less-embeddings,garg2024distributed}.
The basic idea should be obvious: for a real-valued random variable $X$, we know that $ \mathbb{E} [X^{-1}] \ \neq\  \big(\mathbb{E}[X]\big)^{-1}$.
A similar phenomenon occurs for random sketch matrices, $\tilde\A =
\S\A$:
\begin{equation} 
\E[(\tilde\A^\top\tilde\A)^{-1}]\neq (\A^\top\A)^{-1},\quad\text{even
  though}\quad \E[\tilde\A^\top\tilde\A]=\A^\top\A . 
\end{equation} 
This bias is especially pronounced in the proportional regime, due to the random singular value fluctuations described by the MP law.

\vspace{1mm}\noindent\textbf{Why focus on the inverse?}
Let us for a moment consider a statistical perspective on sketching. Here,
the sketched covariance matrix $A^\top  S^\top  S  A$ can be
viewed as a \textit{sample covariance estimator} of the ``population
covariance matrix'' $ A^\top  A \in \mathbb{R}^{d \times d}$. 
A natural question is: how does the spectrum differ between the \textit{sample} and the \textit{population} covariance?
RMT answers this by looking at the \textit{resolvent matrix}: 
$$
( A^\top  S^\top  S  A - z  I)^{-1} \quad \mbox{for} \quad  z \in \mathbb C \setminus \mathbb{R}_+.
$$
The Stieltjes transform (normalized trace of the resolvent, central to RMT) exhibits an \textit{inversion bias}, leading to a discrepancy between the sample and population. 
In traditional RMT, controlling this inversion bias lets us characterize
the limiting eigenvalue distribution as $s,n,d \to \infty$.
Recent works \cite{less-embeddings,precise-expressions,newton-less} have extended
this RMT analysis to the \emph{non-asymptotic}
RandNLA theory, essentially showing that:
  \emph{If we can correct the inversion
   bias of the sketch $\tilde\A=\S\A$, then we can recover stronger Gaussian-like sketching performance for faster sketches across RandNLA tasks.}

\vspace{1mm}\noindent\textbf{Correcting the bias (for Gaussian sketches).}
When $\tilde\A=\S\A$ is an $s\times d$ Gaussian sketch, scaled so that $\E[\tilde\A^\top\tilde\A]=\A^\top\A$, then there is a very simple correction for the inversion bias:
\begin{align*}
  \E\big[(\gamma\tilde\A^\top\tilde\A)^{-1}\big] =
  (\A^\top\A)^{-1}\quad\text{for}\quad
  \gamma= \frac{s}{s-d-1}.
\end{align*}
This simple fix does \emph{not} hold for other (faster) sketching methods that have been used in RandNLA, e.g., Hadamard-based projections, sub-Gaussian sketches, very sparse sketches, sampling methods, etc.
The basic reason is that other sketches are \textit{not} perfectly rotationally symmetric.
Thus, among other things, they could lose rank (with very small probability) and/or suffer from ``coupon collector'' problems.
In general, for faster sketching methods, the \textit{inversion bias occurs differently in each direction}, and so we cannot correct it with a single dimensional rescaling factor.

\vspace{1mm}\noindent\textbf{Nearly-unbiased sketches.}
Going beyond Gaussian sketches motivates the notion of a
nearly-unbiased estimator \cite{less-embeddings}.

\begin{definition}
\label{def:nearly-unbiased-estimator}
A random PSD~matrix $\tilde C$ is an \emph{$(\epsilon,\delta)$-unbiased estimator} of $ C$ if there is an event $\mathcal E$ that holds with probability $1-\delta$ such that,
when conditioned on $\mathcal E$,
$$
\mathbb{E}_{\mathcal E}[\tilde C] \approx_{\epsilon} C \quad\mbox{ and }\quad \tilde C\preceq O(1)\cdot C .
$$ 
\end{definition}

Let $S$ be an $s\times n$ random matrix such that $\sqrt s\, S$ has
i.i.d.\ $O (1)$-sub-Gaussian entries with mean zero and unit
variance (i.e., a direct extension of a Gaussian sketch). Building on the Stieltjes transform analysis, it is
possible to show that for this sketching matrix we can nearly correct
inversion bias, in the sense that $(\frac s{s-d} A^\top S^\top S
A)^{-1}$ is an $(\epsilon,\delta)$-unbiased estimator of $( A^\top
A)^{-1}$ for $\epsilon = O(\frac{\sqrt d}s)$ and $\delta =
e^{-cs}$ (Proposition 4 in \cite{less-embeddings}). In other words, conditioned on a high probability event
$\Ec$, we have
\begin{align}
  \mathbb{E}_{\mathcal E}\big[(\tfrac s{s-d} A^\top S^\top S A)^{-1}\big]
  \approx_\epsilon ( A^\top A)^{-1},
  \quad\text{for}\quad \epsilon = O\bigg(\frac{\sqrt d}s\bigg).\label{prop:sub-gaussian}
\end{align}
\noindent
Let us compare/contrast this notion of an $(\epsilon,\delta)$-unbiased estimator with JL / subspace embeddings in Def.~\ref{def:subspace_embedding}.
For a subspace embedding, we have that
\begin{align*}
  \text{Subspace embedding:}\quad
  ( A^\top S^\top S A)^{-1}
  &\approx_{\eta} ( A^\top A)^{-1},\quad\eta=\Theta\bigg(\sqrt{\frac ds}\bigg).
\end{align*}
Thus, we see that the \emph{average-case} analysis in \eqref{prop:sub-gaussian} is sharper than what is possible to recover with the \emph{high-probability} subspace embedding analysis, by at least a $\sqrt s$ factor.

\vspace{1mm}\noindent\textbf{Averaging RandNLA estimators.}
Implications of this Modern RMT-style analysis for RandNLA are discussed in later sections.
However, as an immediate consequence, we can establish various ``model averaging'' schemes for RandNLA-based estimators, that can be used in statistical/ML/optimization pipelines. 

The premise behind model averaging is that if we can produce a bias-corrected estimator 
$\tilde\x$, i.e., $\text{Bias}^2(\tilde\x) \ll \text{Var}(\tilde\x)$, then we can boost 
its accuracy by averaging multiple independent copies of this estimator. 
When the estimator relies on some linear functional of the inverse matrix
$(\A^\top\S^\top\S\A)^{-1}$, as in sketched LS \cite{garg2024distributed}
and a number of other RandNLA algorithms \cite{newton-less},
then this follows (with some small additional effort) from
\eqref{prop:sub-gaussian}. For example, for $q$ independent sub-Gaussian sketches $S_i$ of size
  $s\geq O(d+q)$, if we average the LS estimates
  $\tilde\x_i = \argmin_\x\|\S_i(\A\x-\b)\|_2$, then the aggregate
  estimator $\bar\x=\frac1q\sum_{i=1}^q\tilde\x_i$ satisfies \cite{garg2024distributed}:
  \begin{align*}
   \E \|\A\bar\x - \b\|_2 \leq
    (1+\epsilon)\|\A\x^*-\b\|_2,\quad\text{for}\quad\epsilon
    = O\Big(\frac d{qs}\Big).
  \end{align*}

\vspace{1mm}\noindent\textbf{Extending RMT-style analysis to fast sketching.}
\michael{As much as we love it, this is an ovbious thing to comment out for the original KDD submission, or move to an appendix for that submission.}
Compared to classical JL or subspace embedding approaches, most RMT for sketching requires
different ``Gaussianization'' assumptions
and different parameter regimes (e.g., the proportional regime).
Most out-of-the-box theory applies only to (expensive) dense Gaussian or (still expensive) dense sub-Gaussian sketching matrices.
Given the widespread interest in fast sketching methods, a natural question is:
\emph{Can we extend this line of work to fast sketches, e.g., sparse or structured?}

Several recent results have provided an affirmative answer to this
question
\cite{DobLiu18_TR,lacotte2020optimal,less-embeddings,gaussianization,chenakkod2023optimal},
including: non-asymptotic RMT-style analysis for sparse sketches based on
 Leverage Score Sparsification (LESS)
 \cite{less-embeddings,gaussianization,chenakkod2023optimal} and
asymptotic RMT-style analysis for structured sketches based on randomized
Hadamard transforms \cite{DobLiu18_TR,lacotte2020optimal}.
These results essentially show that certain fast sketching operators
are ``close'' in a strong enough sense to sub-Gaussian matrices,
typically by relying on two structural conditions which are needed
for $S$ to ensure small inversion bias: 
\begin{enumerate}
  \item \emph{Subspace embedding}:
        this is the standard guarantee from
        Def.~\ref{def:subspace_embedding}; using Modern
        RMT \cite{brailovskaya2022universality}, sharp MP-style
        subspace embedding guarantees as in
        \eqref{eq:mp-law} can be recovered
        for fast sketching operators  such as LESS \cite{chenakkod2023optimal}.
  \item \emph{Restricted Bai-Silverstein inequality}: 
        this is the key novelty that provides a variance bound for random quadratic forms
        \cite{less-embeddings}; and 
        it is related to the Hanson-Wright inequality \cite{rudelson2013hanson}.
\end{enumerate}

\noindent
Based on these ideas, we can reduce the cost of applying sub-Gaussian
sketches down from $O(nds)$, while still recovering \eqref{prop:sub-gaussian},
i.e., we can construct more efficient sketches for which it is possible to
correct inversion bias and recover other Gaussian RandNLA
guarantees. For example, given a tall $n\times d$ matrix $\A$, we can  
  compute an $s\times d$ sketch $\S\A$ (LESS embedding) in near-linear
  time $\tilde O(nd+sd^2)$ such that $(\frac s{s-d} A^\top S^\top S
A)^{-1}$ is an $(\epsilon,\delta)$-unbiased estimator of $( A^\top
A)^{-1}$ with $\epsilon = O(\frac{\sqrt d}s)$, matching the guarantee
for sub-Gaussian sketches (Theorem 8 in \cite{less-embeddings}).

\subsection{RMT for Sampling via DPPs}
\label{sxn:modern-randnla-DPP}

The astute reader may wonder whether the developments described in Sec.~\ref{sxn:modern-randnla-RMT} that hold for data-oblivious sketching methods also hold for data-aware sampling methods.
After all, they are based on RMT-based sketching in the proportional
regime, and in some sense sampling is inherently ``non-RMT.''  
For example, it involves coordinate axes  and coordinate-aligned subspaces, and lower bounds arise that are due to the Coupon Collector problem.
Yet, using Determinantal Point Processes (DPPs), one can show that
very similar RMT results hold for certain data-aware sampling methods \cite{dpps-in-randnla}.

DPPs form a family of non-i.i.d.~sampling distributions with a long history in applied mathematics 
\cite{dpp-physics,  BO00,Bor11,dpp-independence}
that has received interest in ML due to their associations with
diversity and volume \cite{dpp-ml}. In this context, one can define a
DPP for a given $n\times d$ matrix $\A$ as a probability
distribution over the row subsets of $\A$: any subset $S\subseteq
\{1,...,n\}$ is assigned a probability
\begin{align}
\Pr(S) \propto \det(\A_S\A_S^\top),\label{eq:dpp}
\end{align}
where $\A_S$ is the submatrix consisting of the rows of $\A$ indexed
by $S$, and $\det(\A_S\A_S^\top)$ corresponds precisely to the squared
$|S|$-dimensional volume spanned by those rows. Such a distribution
naturally promotes the row subsets that capture more ``information''
about the matrix $\A$, while minimizing the subset size.
This translates to guarantees for Classical
RandNLA tasks which match, and sometimes improve on, those based on
Gaussian (or Gaussian-like) sketching. For examples, it has been shown
 \cite{pca-volume-sampling,more-efficient-volume-sampling,nystrom-multiple-descent} that when $S$ is drawn according to \eqref{eq:dpp} with subset
size $s$, then it achieves a LR approximation guarantee that
matches that of a Gaussian sketch:
\begin{align*}
  \E\,\|\A-\Q\Q^\top\A\|_F^2\leq \Big(1 + \frac{k}{s-k+1}\Big)\cdot\|\A-\A_k\|_F^2,\qquad\Q= \texttt{orth}( A_S).
\end{align*}
This remarkable connection between DPPs and RandNLA has been exploited
not only in LR approximation
\cite{pca-volume-sampling,more-efficient-volume-sampling,nystrom-multiple-descent},
but also in LS regression
\cite{unbiased-estimates,leveraged-volume-sampling} and
iterative optimization
\cite{determinantal-averaging,randomized-newton,derezinski2023solving},
including inversion bias corrections similar
to those that are possible for Gaussian and LESS sketches \cite{unbiased-estimates,debiasing-second-order};
see \cite{dpps-in-randnla} for a detailed discussion.
One way to explain this success is to view DPPs as a natural non-i.i.d. extension of leverage score
sampling. Specifically, it can be shown that the marginal probability
of the $i^{\mathrm{th}}$ row of $\A$ appearing in a DPP sampled set $S$, i.e.,
$\Pr(i\in S)$,
corresponds in a precise sense to the ridge-regularized leverage score
of that row \cite{dpp-intermediate,dpp-sublinear}. These ridge leverage scores have been introduced
(independently of DPPs) as the right sub-sampling technique for
LR approximation and related tasks in a long series of RandNLA works, e.g., \cite{ridge-leverage-scores,musco2017recursive,NIPS2018_7810}.

These ideas have led to (theoretically) fast DPP sampling:
from early approaches for exact sampling via eigendecomposition
\cite{dpp-independence,k-dpp,dpp-tree}; through
so-called intermediate sampling via leverage scores \cite{dpp-intermediate,dpp-sublinear,alpha-dpp}; to
Markov chain Monte Carlo sampling
\cite{rayleigh-mcmc,isotropy-log-concave,anari2022domain,alv22}. Many
of these algorithms are available in the Python package DPPy \cite{dppy}.

Nevertheless, there is still an inevitable computational overhead in
using non-i.i.d.~sampling methods, such as DPPs, in place of
popular i.i.d. sampling techniques such as leverage score and row-norm
sampling. This trade-off is justified when the quality and small size
of the target sample is of paramount importance, e.g., in active
learning, Bayesian optimization, clustering, or coreset
selection. However, the RMT techniques for
analyzing DPPs have also found applications in developing new
guarantees for i.i.d. sampling. In particular, it has been
shown recently \cite{derezinski2023solving,derezinski2024fine} that after preprocessing matrix $\A$ with a randomized Hadamard
transform, an i.i.d. uniform sample of $O(k\log k)$ rows with high probability contains
a DPP sample of size $k$. This has unlocked new approaches in
using RandNLA sampling methods for stochastic optimization methods
such as Sketch-and-Project (see Sec.~\ref{sec:optimization}).

\section{Advances in RandNLA for Optimization}
\label{sec:optimization}

In stochastic optimization, the goal is to minimize (or maximize) an
objective function, when one or more of the input parameters is
subject to randomness. It arises in many areas, most prominently with
stochastic gradient descent (SGD) based methods for training ML models
\cite{large-scale-tradeoffs}. Leveraging Modern RandNLA techniques and
theory (from Sec.~\ref{sxn:modern-randnla}) within the framework of stochastic optimization leads to the
Iterative Sketching paradigm (of Fig.~\ref{fig:schematic-three-paradigms}).
This has manifested
itself in many ways, including: PW-SGD \cite{yang2017weighted}, a
variant of SGD which is preconditioned and weighted using RandNLA
techniques; Subsampled Newton
\cite{erdogdu2015convergence,subsampled-newton,XYRRM16_TR} and Newton
Sketch \cite{pilanci2017newton}, which apply the Sketch-and-Solve
paradigm to Hessian estimation; and Sketch-and-Project
\cite{gower2015randomized,derezinski2022sharp,derezinski2023solving},
a modern extension of the classical Kaczmarz method \cite{kaczmarz37},
which uses sketching to achieve \textit{implicit preconditioning} in stochastic methods.
Recent advances in RandNLA (described in
Sec.~\ref{sxn:modern-randnla-RMT}) have proven fruitful in obtaining
theory for stochastic optimization algorithms based on Iterative
Sketching, which comes with much smaller theory-practice gap compared to
traditional approaches. 

Consider the following standard finite-sum optimization task:
\vspace{-3mm}
\begin{align}
    \text{Minimize}\qquad f(\x) = \frac1n\sum_{i=1}^n\psi_i(\x)  ,
    \label{eq:finite-sum}
\end{align}
\vspace{-3mm}

\noindent
where $\x$ is a parameter (or weight) vector, and
where each $\psi_i(\x)$ corresponds to the loss for a single data point. 
In optimization, we generally approach this problem by iteratively refining a sequence of estimates $\x_0,\x_1,\x_2,...$, relying on local information about the objective function. 
We can divide optimization methods into \emph{first-order} methods,
which use gradient (first derivative) information of $f$ at $\x_t$,
e.g., gradient descent; and \emph{second-order} methods, which
additionally rely on Hessian (second derivative) information. 

Although introducing randomization in optimization algorithms has
proven very effective in ML, with variants of SGD (AdaGrad, Adam,
SVRG, etc.) being widely-used in practice, SGD-based methods still
often suffer from instability, large variances, and extreme
sensitivity to hyperparameter choice
\cite{yao2021adahessian,HodMah20A_TR}. 
In this context, randomized algorithms based on Modern RandNLA theory
can be used to provide better stability, to inject better curvature
information, and to reduce communication and computation costs.

\subsection{Gradient Sketch}
\label{sec:sgd}

With SGD, we estimate the first-order (gradient) information by subsampling the components $\psi_i$ of the objective, and using those to construct an unbiased estimate $\hat\g_t$:
    \begin{align*}
      \x_{t+1} = \x_t - \eta_t\hat\g_t,\qquad\text{where}\quad
      \E[\hat\g_t] = \nabla f(\x_t).
    \end{align*}
Despite its popularity, SGD has a number of limitations, e.g., large variance $\E\big[\|\hat\g_t-\g_t\|^2\big]$, which slows the convergence near the optimum, as well as
sensitivity to hyper-parameters (such as step size, mini-batch size, etc.). 
RandNLA offers a number of techniques for addressing these
limitations, such as using weighted gradient sampling based on
  leverage scores or other importance scores, as well as
  using a sketching-based preconditioner.
We consider the following.

\vspace{1mm}\noindent\textbf{Preconditioned Weighted SGD.}
Consider the Preconditioned Weighted SGD (PW-SGD) method
(which is shown here for the LS problem, $f(\x)= \frac1n\|\A\x-\b\|^2$, but which is applicable far beyond this setting \cite{durfee2018ell_1,agarwal2017second,gonen2016solving,svrn}):  

   \begin{algorithmic}[1]
    \State Compute $SA$ with some sketching operator $S$
    \State Compute $R$ such that  $SA =QR^{-1}$ for orthogonal $Q$
    \State Compute leverage score estimates $\tilde l_i$ for $A$
    \For{$t=0$ to $T-1$}
    \State Compute $g_t \leftarrow \frac1s\sum_{i=1}^s\frac Z{\tilde l_{I_i}}\nabla \psi_{I_i}(\x_t),\quad \Pr(I_i)\propto  \tilde l_{I_i}.$
    \State Compute $x_{t+1}\leftarrow x_t - \eta_t RR^\top g_t$ 
    \EndFor
  \end{algorithmic}
This algorithm uses a sketching operator $S$ to construct the $R$, which is the preconditioner of the problem, since its spectrum approximates the spectrum of (the inverse of) $\A$. 
Moreover, the algorithm uses leverage score estimates of $\A$,   
which are used
for
subsampling the gradients (but this can be replaced by other sketching/sub-sampling schemes).

Thanks to a combination of preconditioning and importance sampling,
this version of PW-SGD can completely avoid any condition number
dependence in its convergence rate \cite{chenakkod2023optimal}.
In particular, 
    suppose that the \textit{leverage score} estimates satisfy:
    $\tilde l_i\geq l_i(\A)/\alpha$ for all $i$.
    Then, letting $\eta_t:= \frac{\beta}{1+\beta t/8}$ for 
    $\beta = \frac{k/8}{k+4\alpha d}$, this version of PW-SGD 
    satisfies: 
    \begin{align*}
      \E[f(x_t) - f(x^*)] \leq \frac{f(x_0)}{1+st/(c\alpha d)}\qquad\forall_{t\geq 1}. 
    \end{align*}

\noindent
The key advantage here is that the resulting iteration complexity $t=O(d/s\epsilon)$
is entirely independent of the number of data points $n$ or of the
condition number $\kappa$ of matrix $\A$ \cite{chenakkod2023optimal},
whereas classical SGD may require as many as $O(n\kappa^2/s\epsilon)$
iterations. 
For reaching a \textit{moderate precision} solution (the regime of greatest interest in ML and data science applications), the computational cost comparison shows that this method is faster than Sketch-and-Solve or Sketch-and-Precondition \cite{chenakkod2023optimal}.

There are also other RandNLA-based approaches for sketching the gradient
information of the objective: 
in non-finite-sum settings, e.g., Sega \cite{hanzely2018sega};
for distributed/federated learning, e.g., FetchSGD \cite{rothchild2020fetchsgd}; 
methods inspired by randomized coordinate descent \cite{needell2014stochastic,lee2013efficient};
and randomized preconditioning for other stochastic gradient methods,
e.g., Preconditioned SVRG and SVRN \cite{gonen2016solving,svrn}, and others \cite{gower2018tracking,liu2019acceleration}.

\subsection{Hessian Sketch}

RandNLA methods have proven particularly effective at efficiently extracting second-order information about the optimization objective \cite{subsampled-newton,pilanci2017newton,newton-less}. 
This has led to many sketching-based Newton-type optimization algorithms. 
Recall that Newton's method represents the paradigmic second-order optimization algorithm, which minimizes a local quadratic approximation of the objective using gradient $\g_t=\nabla f(\x_t)$ and Hessian $\H_t=\nabla^2 f(\x_t)$:
\begin{align*}
  \x_{t+1} = \x_t + \argmin_\v\Big\{\g_t^\top\v +
  \frac{\eta_t}2\v^\top\H_t\v\Big\} = \x_t-\eta_t\H_t^{-1}\g_t  .
\end{align*}

\noindent
As an example, consider a standard Generalized Linear Model~(GLM),
    $f(\x) = \frac1n\sum_{i=1}^nl_i(\a_i^\top\x) +
    \frac\gamma 2\|\x\|^2$,
  where $\a_i^\top\x$ represents a linear prediction associated with a data
  point $\a_i\in\R^d$, and loss $l_i$ encodes the prediction error, dependent
  on a label $y_i$. For instance, in logistic regression, we have $y_i=\pm 1$
  and $l_i(\a_i^\top\x) = \log(1 + e^{-y_i\a_i^\top\x})$. Here, the
  Hessian at $\x_t$ is given by:
    \begin{align*}
      \nabla^2 f(\x_t) = \frac1n\A^\top\D_t\A+\gamma\I,
      \qquad \D_t := \diag(l_1''(\a_1^\top\x_t),...,l_n''(\a_n^\top\x_t)),
    \end{align*}
where the dominant cost is the $O(nd^2)$
matrix multiplication $\A^\top\D_t\A$. 
It is natural to reduce this cost with
RandNLA sketching or sub-sampling, by repacing $\A^\top\D_t\A$ with
$\tilde\A_t^\top\tilde\A_t$ where $\tilde\A_t=\S_t\D_t^{1/2}\A$ for some
sketching matrix $\S_t$.
One version of this leads to the
Newton Sketch \cite{pilanci2017newton}:
\begin{align*}
  \x_{t+1} 
   = \x_t-\eta_t \hat\H_t^{-1}\g_t,\quad \hat\H_t = \frac1n\tilde\A_t^\top\tilde\A_t+\gamma\I.
\end{align*}
Standard RandNLA guarantees such as the subspace embedding are
sufficient (although not necessary) to ensure that $\hat \H_t$ provides a good enough
approximation to enable accelerated local convergence in time $\tilde O(nd)$.

These approaches have also been extended to distributed settings via
RMT-based model averaging, 
with
applications in ensemble methods, distributed optimization, and federated learning
\cite{lacotte2022adaptive,lacotte2020optimal,hessian-averaging,newton-less,gupta2021localnewton}.  
Further RandNLA-based Newton-type methods include:
Subsampled Newton
\cite{erdogdu2015convergence,subsampled-newton,bollapragada2018exact,berahas2020investigation};
Hessian approximations via randomized Taylor expansion
\cite{agarwal2017second} and low-rank approximation \cite{frangella2023promise,derezinski2024faster}; Hessian
diagonal/trace estimates via Hutchinson's
method \cite{meyer2021hutch++} and Stochastic Lanczos Quadrature, particularly for
non-convex problems,
e.g., PyHessian \cite{yao2020pyhessian}, AdaHessian
\cite{yao2021adahessian}; and finally
Stochastic Quasi-Newton type methods \cite{kovalev2019stochastic,mokhtari2018iqn}.

\subsection{Sketch-and-Project}

The Sketch-and-Project framework has gained
attention as a powerful methodology within the Iterative Sketching
paradigm of RandNLA. While this framework has found applications across
the stochastic optimization landscape, it originally arose from
randomized algorithms for solving systems of linear
equations.  

Solving an $m\times n$ linear system $\A\x=\b$ can be viewed as an instance of the LS problem, i.e., minimizing the objective $f(\x) = \|\A\x-\b\|_2^2$. 
As discussed in Sec.~\ref{sxn:least-squares}, Classical RandNLA has addressed this problem in the \emph{highly} over- or under-determined settings (i.e., when $\A$ is very tall or wide).
However, in many applications, particularly high-dimensional
settings that arise in ML,
$m$ and $n$ are (equal or) of comparable size.

For these nearly-square matrix problems, randomization can still be beneficial, in particular when using Modern RandNLA methods. 
For instance, consider the classical Kaczmarz algorithm, which solves
a linear system of $m$ equations $\a_i^\top\x=b_i$ via the following
iterative procedure, starting from some $\x_0$:
\begin{algorithmic}[1]
    \For{$t=0$ to $T-1$}
    \State Select index $I_t\in\{1,...,m\}$
    \State $\x_{t+1}\leftarrow$ Project $\x_t$ onto the solutions of equation $\a_{I_t}^\top\x=b_{I_t}$
    \EndFor
  \end{algorithmic}
This simple procedure has been known for a long time \cite{kaczmarz37}, but only with the use of randomization are we able to characterize its convergence \cite{strohmer2009randomized}:
    if the Kaczmarz method selects index $I_t$ randomly, with probability proportional to $\|\a_{I_t}\|_2^2$, then:
    \begin{align*}
      \E\,\|\x_t - \x^*\|_2^2 \leq \bigg(1 - \frac{\sigma_{\min}^2(\A)}{\|\A\|_F^2}\bigg)^t\|\x_0-\x^*\|_2^2.
    \end{align*}
  We can interpret this Randomized Kaczmarz (RK) method as a Weighted SGD algorithm 
  solving the finite-sum minimization problem \eqref{eq:finite-sum} with
  $\psi_i(\x)=(\a_i^\top\x-b_i)^2$ \cite{needell2014stochastic}, 
  drawing a parallel with PW-SGD as described in Sec.~\ref{sec:sgd}. 

However, this paradigm becomes quite different from PW-SGD once we
select more than one equation at a time, giving rise to
Sketch-and-Project \cite{gower2015randomized}: 
\emph{Sample a random $k\times m$ matrix $S=S(t)$,
  and project $\x_t$ onto the solutions of $\S\A\x=\S\b$:}
  \begin{align*}
    \x_{t+1} = \argmin_\x\|\x_t-\x\|_2^2\quad\text{subject to}\quad \S\A\x=\S\b.
  \end{align*}
We recover RK if the matrix $\S$ is chosen to be the indicator vector of the equation $I_t$.
However, this general framework (with Modern RandNLA tools from Sec.~\ref{sxn:modern-randnla}) allows for other natural choices,
such as selecting blocks of equations (Block Kaczmarz, e.g., \cite{needell2013two,needell2014paved,rebrova2021block}) 
or sketching the input matrix $\A$ using any of the methods described earlier. 
The Sketch-and-Project framework has been used to capture and extend other stochastic optimization methods, including Coordinate Descent
\cite{leventhal2010randomized, lee2013efficient, gower2018accelerated}, and to develop more general purpose
first- and second-order optimization algorithms \cite{gower2019rsn,kovalev2019stochastic,hanzely2020stochastic,yuan2022sketched}.
      
Going beyond RK, the convergence analysis of Sketch-and-Project has proven much more challenging, largely due to the complex interdependence between the distribution of the sketching matrix $S$ and the spectrum of the input matrix $\A$. 
However, by using tools from Modern RandNLA theory \cite{rk20,randomized-newton,precise-expressions},
we can relate this convergence to the quality of the sketch $\S\A$ as
a low-rank approximation of the data \cite{derezinski2022sharp}:
    \begin{align}
      \E\,\|\x_t-\x^*\|_2^2 \ \lesssim\  \bigg(1 - \frac{k\sigma_{\min}^2(\A)}
      {\E\|\A- \A
      \P_{\S\A}\|_F^2}\bigg)^t\,\|\x_0-\x^*\|_2^2.\label{eq:implicit-preconditioning}
    \end{align}
Here, $\E\|\A - \A \P_{\S\A}\|_F^2$ is the projection error of
the low-rank approximation of $\A$ produced from the sketch $\S\A$,
same as in equation \eqref{eqn:LR_structural_result} from  Sec.~\ref{sxn:low-rank} (except sketched on
the other side). This analysis has revealed new classes of problems
(often overlapping with ML domains that exhibit low-rank structure)
for which Sketch-and-Project methods offer dramatically improved peformance over
the more traditional Sketch-and-Precondition paradigm
\cite{derezinski2023solving,derezinski2024fine}, thanks to the 
\textit{implicit preconditioning} phenomenon that is described by \eqref{eq:implicit-preconditioning}.

\section{Advances in RandNLA for ML and Statistics}
\label{sec:statistics}

So far, we have primarily focused on how RandNLA can be used to
address ML and data science problems from a purely algorithmic perspective:
the data (e.g., matrix $\A$ and vector $\b$) is given and deterministic, and our
goal is to directly compute or estimate a property of the data (e.g.,
the LS solution). With this problem formulation,
randomization has entered solely from the algorithm design perspective,
and thus it is fully controllable by the practitioner.

However, when we
consider the same problems from a statistical 
perspective, the data is often assumed to be random itself, e.g.,
coming from some data distribution (which may or may not be known) and
distorted by some random noise (e.g., Gaussian). In this case, our goal
is often to estimate a quantity that may depend on an unobserved part
of the data distribution, in which case we must account for the
\emph{generalization error}, in addition to the \emph{approximation error}
that was our focus so far. In this section, we delve into how
randomization coming from RandNLA interacts with the inherent data
randomness, and how that affects the design and implementation of
these methods.

\subsection{Statistical Learning Approaches}
\label{sec:statistics-learning}

We start by considering a statistical learning theory perspective on
generalization error, which is most traditionally exemplified by the
probably approximately correct (PAC) model of learning. Here, we
often assume very little about the underlying data distribution,
leading to a worst-case view of the generalization error. This forces
our algorithmic approaches to exhibit a certain degree of robustness
to the unobserved data. We show that RandNLA techniques naturally
inject such robustness into ML algorithms, as they are also designed
to work well under worst-case settings. 
  
\paragraph{Robust semi-supervised learning.}
As a key motivating example, let us return to LS regression. So
far, we have framed this problem so that given a matrix $\A$ and vector
$\b$, we want to find $\x^*=\argmin_\x\|\A\x-\b\|_2^2$. However, what if part
of this data is withheld, for instance due to the cost of procuring
the labels?
  
Formally, suppose that we are given a data matrix $\A\in\R^{n\times d}$ with $n$ unlabeled data points:
  $\a_1,\dots,\a_n\in\R^d$ (transposed rows of $\A$). For each point $a_i$, we can decide to
  query the random label $b_i\in\R$. Our goal is to query subset $S\subseteq\{1,...,n\}$
  of at most $s\ll n$ labels and output $\hat\x$ that has small error
\begin{align*}
\E\,f(\hat\x):=\E\,\frac1n\sum_{i=1}^n(\a_i^\top\hat x- b_i)^2 = \E\, \frac1n\|\A\hat\x-\b\|_2^2.
\end{align*}
Note that, due to the underlying statistical randomness and the fact
that we do not have access to all of the data, this is now a
generalization error, and the problem is not purely computational in nature.
The question is: \textit{How should we choose the data points to label?}

This task might seem dramatically different than the classical LS task that is tackled by Classical RandNLA. 
Yet, recent works have shown that RandNLA subsampling techniques, such as leverage
score sampling and DPPs, play a crucial role in ensuring good
generalization error in this setting. In particular, it is known (e.g.,
\cite{Mah-mat-rev_BOOK,leveraged-volume-sampling}) that
    for any fixed $\A$ and random $\b$,  if we query a subset $S$ of  $s=O(d\log
    d+d/\epsilon)$ entries of $\b$ sampled according to $\A$'s
    leverage scores, and compute $\hat\x = \argmin_\x\|\A_S\x-\b_S\|_2^2$, then:
    \begin{align*}
      \E\,f(\hat\x) \leq (1+\epsilon) \min_\x\E\,f(\x).
    \end{align*}
  Crucially, the sampling method used here does not require any
  knowledge about the label distribution, which means that it gives a
  very robust generalization error guarantee. In fact, one can show
  that randomization is actually necessary in the label selection, to
  protect against adversarial label distributions \cite{unbiased-estimates-journal}.

  This robust learning via RandNLA sampling has been further refined
  by relying on non-i.i.d. approaches such as DPPs (using tools from Sec.~\ref{sxn:modern-randnla-DPP}) and related methods
  \cite{unbiased-estimates,leveraged-volume-sampling,chen2017condition}. It
  has also been extended to other regression losses via different
  sampling schemes     (inverse scores, Lewis weights, ridge leverage,
  sensitivity scores, etc.)
  \cite{minimax-experimental-design,chen2021query,parulekar2021l1,musco2022active,chen2022online},
  as well as to other statistical learning settings such as single index
  models \cite{gajjar2024agnostic} and generalized linear models \cite{jambulapati2023sparsifying}.

  \paragraph{Kernel-based learning.}
  RandNLA subsampling methods have also been extended to ``non-linear''
learning tasks, via kernel-based learning methods. These approaches
work by constructing a mapping $\phi:\Xc\rightarrow \Fc$ from the data space $\Xc$ to an
expanded feature space $\Fc$. While the expanded data vectors are never
explicitly computed, we can learn about the geometry of the dataset by
looking at the pairwise inner products between $i^{\mathrm{th}}$ and $j^{\mathrm{th}}$ data
point, which are obtained through the so-called kernel function
$k(\x_i,\x_j)=\phi(\x_i)^\top\phi(\x_j)$. In the resulting learning
problem, we receive a dataset
$(\x_i,y_i)_{i=1}^n\sim\Dc$, and construct an estimator $\tilde f$ for $f^*$ by minimizing a
regularized objective over the kernel space $\Fc$:
\begin{align*}
 \min_{f\in\Fc}\ \frac1n\sum_{i=1}^n\ell(f(\x_i),y_i) + \frac\lambda 2\|f\|_{\Fc}^2,
\end{align*}
where $\|f\|_{\Fc}$ is the norm induced by $\Fc$ and $\ell$ is a
loss function. Using square loss and the representer theorem, this
can be cast as kernel ridge regression (KRR), which is equivalent to solving the following regularized linear system:
\begin{align}
  (\K + n\lambda\I)\alpha = \y,\label{eq:krr}
\end{align}
where $\K=\big[\phi(\x^{(i)})^\top\phi(\x^{(j)})\big]_{ij}$ is the
$n\times n$ PSD kernel matrix that includes all
pairwise inner products in the feature space, and $\alpha$ is a vector
from which we can recover the prediction function $f$ as
$f(\x_i)=[\K\alpha]_i$. 

Naturally, we could now use RandNLA techniques, such as those
discussed in Sec.~\ref{sec:optimization}, to solve
\eqref{eq:krr} directly. However, this being a statistical learning
problem significantly affects our algorithmic choices. To start, note
that we are not given $\K$ as input, but rather we must compute each
of its entries via the kernel function $k(\cdot,\cdot)$. In fact, even
just forming $\K$ at the cost of $n^2$ kernel evaluations is a
computational bottleneck we may wish to avoid. On the other hand, due
to the presence of generalization error, we do not need to solve the
problem to high precision.

We can address both of these considerations by exploiting the fact
that the kernel matrix $\K$ has a PSD 
structure. This allows us to approximate it efficiently without even
forming the entire matrix $\K$ by relying on a \emph{Nystr\"om approximation}:
 \begin{align*}
     \Tilde\K  &= \K\S(\S^\top\K\S)^{-1}\S^\top\K,
 \end{align*}
 where $\S$ is an  $n\times s$ subsampling matrix such that $\K\S$
 selects $s$ columns from $\K$ and $\S^\top\K\S$ selects an $s\times
 s$ principal submatrix of $\K$. 
Thus, the Nystr\"om method requires only $ns$
kernel evaluations, i.e., \emph{sublinear} in the size of $K$. This approach is closely linked to the LR
approximations described in Sec.~\ref{sxn:low-rank}: using the PSD
property of 
the kernel matrix $K$, we can express it as $K=\A^\top\A$ for some matrix
$\A$; then, the Nystr\"om approximation is given by
$\Tilde\K=\A^\top\P_{\A\S}\A$, and its spectral/Frobenius error is
$\|\K-\tilde\K\|_\xi=\|\A-\P_{\A\S}\A\|_{\xi}^2$, which corresponds to the LR
projection error \eqref{eqn:LR_structural_result} for $\A$.

A long line of works
at the intersection of ML and RandNLA have shown that this method can
be used to produce kernel ridge regression estimators with strong generalization
guarantees faster than it takes to form the kernel matrix $K$. While early efforts proposed using uniform subsampling to
select the rows/columns of $\K$
\cite{Williams01Nystrom,bach2013sharp,rudi2017falkon}, more recent
works have shown that RandNLA-style subsampling using
\emph{ridge leverage scores} yields more robust estimators
\cite{ridge-leverage-scores,musco2017recursive,NIPS2018_7810}. This
has been further improved with DPP sampling
\cite{dpp-sublinear,sparse-variational-gp,nystrom-multiple-descent}. 

Additional related works on the Nystr\"om method include applications to
Gaussian Process regression
\cite{sparse-variational-gp,calandriello_2019_coltgpucb},
Nystr\"om-based sketching
\cite{revisiting-nystrom,balabanov2022block} and preconditioning \cite{avron2017faster,frangella2021randomized,frangella2023promise,derezinski2024faster}, sublinear time LR
approximation of other structured matrices
\cite{musco2017sublinear,bakshi2018sublinear}, 
and sublinear time DPP 
sampling algorithms \cite{dpp-sublinear,alpha-dpp,alv22}.

\subsection{Statistical Inference Approaches}
\label{sec:statistics-inference}

Next, we consider how RandNLA interacts with and informs statistical inference
approaches to modeling and analyzing data. In statistical inference,
it is common to impose strict assumptions on the generative model of
the data, e.g., that the feature vectors are coming from a Gaussian or
sub-Gaussian distribution, and that the predicted variable follows
some underlying linear model distorted by noise.
This generative modeling approach enables a wide range of inferential
tools for designing, optimizing and evaluating estimators. 
These tools include cross-validation, feature selection, reliable
confidence intervals, the Bootstrap method, the Jackknife, etc. This might seem in stark
contrast to the RandNLA approach to data, which has largely been
centered on robustness to the worst-case and avoiding data
assumptions. As we have seen in previous sections, a typical framework
for RandNLA algorithms is that we are given an arbitrary input matrix
$\A$ with no distributional assumptions.

\begin{figure}
  \centering
  \resizebox{0.6\textwidth}{!}{
    \begin{tikzpicture}[scale = 0.7]
      \draw (-3,1.4) node () {\emph{\textbf{Sketching}}};
      \draw (3,1.4) node () {\emph{\textbf{Statistics}}};
  
      \node[fill=blue!5,draw,shape=ellipse,minimum height=1cm,
      minimum width=3cm] (Data) at (-3,0)
    {Big data};    
    \node[fill=blue!5,draw,shape=ellipse,minimum height=1cm,
    minimum width=3cm]
    (Population) at (3,0) {Population};

    \draw (-6,1.25) node (Arbitrary) {\footnotesize Arbitrary input};
    \draw (6,1.25) node (Gaussian) {\footnotesize Gaussian input};
    \draw [->,red!50,line width=1,  shorten >= 1mm]
    (Arbitrary) -- (Data);
    \draw [->,red!50,line width=1,  shorten >= 1mm]
    (Gaussian) -- (Population);

    \node[fill=red!5,draw,shape=ellipse,minimum height=0.75cm]
    (Sketch) at (-3,-2.25) {Sketch};
    \node[fill=red!5,draw,shape=ellipse,minimum height=0.75cm]
    (Sample) at (3,-2.25) {Sample};
    
    \draw [thick,->, shorten >= 1mm]
    (Data) -- (Sketch);
    \draw (-1.8,-1.) node () {\footnotesize Algorithmic};
    \draw (-1.5,-1.35) node () {\footnotesize Gaussianization};

    \draw [thick,->, shorten >= 1mm] 
    (Population) -- (Sample);

    \node[fill=white,draw,shape=rectangle,
    rounded corners=0.15cm,minimum height=6mm, minimum width=3.5cm]
    (Stat) at (0,-4) {Statistical model};

    \draw [thick,->, shorten >= 2mm]
    (Sample) -- (Stat);
    \draw [thick,->, shorten >= 2mm]
    (Sketch) -- (Stat);   
  \end{tikzpicture}}
\vspace{-1mm}\caption{%
  An illustration of the connection between RandNLA and statistical inference.
  Even though the input matrix~$\A$ for a
  RandNLA algorithm is often arbitrary and deterministic, the
  Algorithmic Gaussianization effect of sketching
  turns this matrix into a random data sample~$\tilde \A$ (the sketch),
  which follows a generative model like  those in
  statistical~inference.}\label{fig:inference}
\vspace{-1.5mm}
\end{figure}
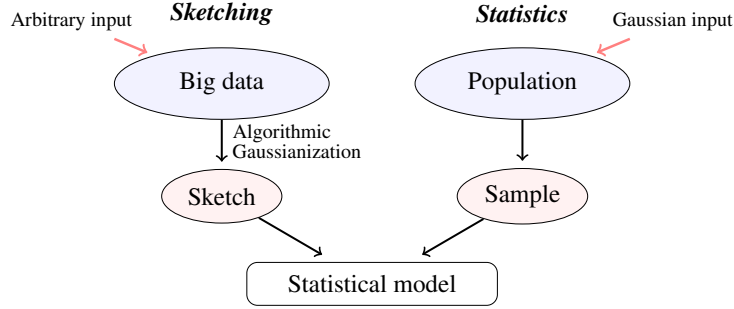

To see that there is a strong connection, however, recall that instead of operating directly on $\A$, we first compute a smaller sketch $\tilde\A=\S\A$, and then we use that sketch to estimate the target properties of the data. 
\emph{This means that, even though the input matrix $\A$ is deterministic and arbitrary, the matrix $\tilde\A$ which we use for the data analysis follows a well-defined generative model---and one which, for many sketching operators, is very close to a sub-Gaussian data model used in statistical inference.} 
See Fig.~\ref{fig:inference} for an illustration. This motivates a statistical view of the
numerical algorithms based on sketching, unlocking the vast wealth of 
inferential tools that can be applied to any approaches
based on the Sketch-and-Solve paradigm and beyond.

\paragraph{Bootstrap error estimation for sketching.}
One application of this viewpoint is error estimation. A common short-coming of
Classical RandNLA error guarantees is that they are overly
pessimistic due to their worst-case nature, and therefore, they do not
provide accurate predictions of the actual error incurred by the
produced estimates. For instance, consider the
sketched LS estimator: given a deterministic matrix
$A\in\R^{n\times d}$ and a vector $b\in\R^n$, reduce the problem with a random sketching matrix $S\in\R^{s\times
  n}$, defining $\tilde A:=SA$ and $\tilde b:=Sb$, and use that to
compute
$\tilde x:=\argmin_\x \|\tilde Ax-\tilde
b\|_2.$ How well
does this particular estimate $\tilde \x$ approximate the optimum
$\x^* := \argmin_\x\|\A\x-\b\|_2$? In other words, we would like
to efficiently compute a \textit{numerical estimate} $\tilde q(\alpha)$ such that
\begin{align*}
  \|\tilde x- \x^*\|_2 \leq \tilde q(\alpha)
  \qquad \text{with
  probability }1-\alpha.
\end{align*}
In statistical inference, this is often done using the Bootstrap
method: by artificially generating another estimate
$\hat x$ such that the fluctuations of $\hat x-\tilde x$ are
similar to the fluctuations of $\tilde x-x^*$. This so-called
bootstrapped estimate $\hat x$ can be obtained as the LS
solution obtained by ``perturbing'' $\tilde A$ and $\tilde b$ (e.g.,
by sub-sampling with replacement). While traditional analysis of the
boostrap has required a sub-Gaussian model of the data, in RandNLA we
can rely on the Gaussianizing effect of sketching and RMT analysis to show that
bootstrap error estimation is provably reliable, even for arbitrary
input matrices $\A$ \cite{lopes2018error,lopes2019bootstrap}.

\paragraph{Separation between sketching methods.}
Another application of the statistical viewpoint on RandNLA is
providing a fine-grained performance analysis of
different sketching or estimation methods which are indistinguishable
under worst-case bounds
\cite{DobLiu18_TR,chi2018randomized,lacotte2020optimal,surrogate-design}. In
particular, such analysis has been 
performed for the above sketched LS estimator, showing that
with a fixed sketch-to-data aspect ratio $s/n$, asymptotic error for 
  orthogonal (Hadamard or Haar) sketches is less than for i.i.d/sub-Gaussian
  sketches \cite{DobLiu18_TR,lacotte2020optimal}.

  Further related advances in applying statistical inference
  techniques to RandNLA include fine-grained analysis of RandNLA
  importance sampling, e.g.,
    \cite{MMY15,GarveshMahoney_JMLR,ma2022asymptotic}, and LR
    approximation  \cite{yang2020reduce}, bootstrapping sketched
    covariance estimation \cite{lopes2023bootstrapping}, as well as
    uncertainty quantification for randomized linear system 
    solvers \cite{bartels2019probabilistic,cockayne2021probabilistic}.

\subsection{Connections with Modern RMT}  
\label{sec:statistics-rmt}

One of the key tools in Modern RandNLA for developing fine-grained understanding of
the performance of RandNLA algorithms, including those inspired by
the above statistical inference viewpoint,  has
been the use of techniques from both asymptotic and non-asymptotic
RMT. These techniques are particularly important when
dealing with matrices with a statistically well-understood
data distribution, where worst-case performance guarantees can be
significantly misleading.

While Classical RandNLA has focused on describing the performance
of the algorithms solely in terms of the dimensions of the input (or
possibly also its sparsity), it has been shown that for many tasks it
is the spectral decay profile of the data (i.e, the rate of descrease
of the singular values of a matrix $\A$) that is most informative
about the performance of these algorithms. In fact, for many ML tasks, the spectral decay
profile of the data can often be sharply characterized, for example,
in terms of whether we expect a heavy-tailed, short-tailed, or spiked profile~\cite{MM18_TR}. 
This can have a dramatic effect on the design and performance of RandNLA algorithms.

\begin{figure}
\centering  \includegraphics[width=0.6\textwidth]{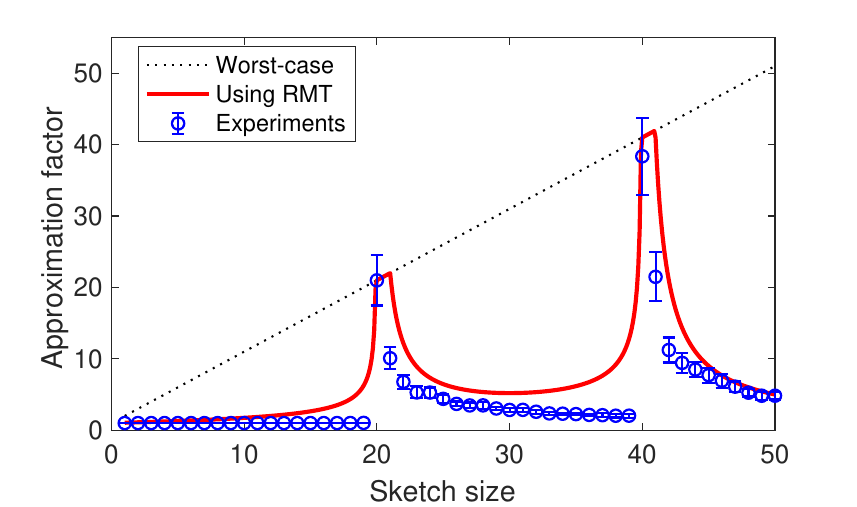}
  \caption{Fine-grained RMT analysis of LR approximation
    reveals a multiple-descent phenomenon that is not predicted by
worst-case analysis.}\label{fig:multiple-descent}
\end{figure}

\paragraph{Multiple-descent in LR approximation.}
As an example, consider the task of LR matrix approximation. As
discussed in Sec.~\ref{sxn:low-rank}, in this task our goal is to
construct a near-optimal rank $k$ approximation of an $m\times n$
matrix $\A$. A simple way of doing this involves
computing an $m\times k$ sketch $\A\S$ and then projecting the matrix $\A$ onto the
rank $k$ subspace defined by $\A\S$. Then, we can measure the quality
of the output by comparing its error to the error of the best rank $k$
approximation of $\A$, obtained via the Top-$k$ Truncated Singular
Value Decomposition (Top-$k$ SVD):
  \begin{align*}
  \text{Low-rank approximation factor}\quad=\quad
  \frac{\|\A-\P_{\A\S}\A\|_F^2}{\|\A-\A_k\|_F^2},
  \end{align*}
  where $\|\A-\P_{\A\S}\A\|_F^2$ is the
  LR projection error as in \eqref{eqn:LR_structural_result}, and
  $\|\A-\A_k\|_F^2$ is the Top-$k$ SVD error. Classical worst-case analysis
  suggests that even for Gaussian-like sketching methods, the
  approximation factor for this basic approach is large and grows
  with $k$, as seen in Fig.~\ref{fig:multiple-descent} (black dotted
  line). This has been
  addressed by computationally more expensive approaches that involve
  increasing the sketch size or using block power
  iteration \cite{tropp2011structure}. However, recent RMT-style
  analysis (Fig.~\ref{fig:multiple-descent}, red solid line; see \cite{nystrom-multiple-descent,precise-expressions})
 reveals that for many data matrices, 
  the true approximation factor is much smaller, i.e., it typically does not
  grow with $k$, and we can characterize it fully based on the
  spectral decay profile of the matrix $\A$. These ideas have been further developed
in other applications of RandNLA, including randomized PCA
\cite{chi2018randomized}, Gaussian Processes \cite{sparse-variational-gp}, iterative
optimization \cite{newton-less,lacotte2020optimal}, and in connections with the double descent phenomenon in over-parameterized ML models \cite{surrogate-design}.

\paragraph{Implicit regularization in sketching.}
Another important consequence of the fine-grained RMT analysis in
RandNLA is a characterization of the \emph{implicit regularization}
phenomenon. Implicit regularization  commonly refers to the
regularizing effect that stochastic algorithms (such as SGD)
have on the trained ML model \cite{Ney17_TR,MM18_TR}; but, 
fundamentally, it has to do with explicit statistical problems that fast approximation algorithms implicitly solve exactly \cite{MO11-implementing,Mah12}. 
Empirical results  \cite{wang2018sketched,debiasing-second-order}
have shown that such a regularizing effect is caused by the stochasticity coming
from the use of sketching operators; and recent theory makes this precise~\cite{surrogate-design,lejeune2020implicit,lejeune2024asymptotics}.

A key example of this phenomenon occurs in sketched ridge regression,
which is an extension of the LS task, where given $\A$,
$\b$ and parameter $\gamma$, we aim to
approximate $\x^* = \argmin\|\A\x-\b\|_2^2 + \gamma\|\x\|_2^2$ by relying
on the sketched data $\S\A$ and $\S\b$. This raises the question of
what regularization parameter $\tilde\gamma$ should be used when computing the
sketched estimate:
\begin{align*}
  \tilde\x = \argmin_\x\|\S(\A\x-\b)\|_2^2+\tilde \gamma\|\x\|_2^2.
\end{align*}
Here, again, Classical RandNLA analysis is not sensitive enough to
detect the implicit regularization effect, and suggests that the correct
regularization parameter should be the same as for the original task, $\tilde\gamma=\gamma$. However,
recent fine-grained results have enabled characterizing the bias of
the sketched estimators showing that we must shrink the
regularization parameter $\tilde\gamma$ to correct for implicit
regularization \cite{debiasing-second-order}:
\begin{align*}
  \E[\tilde\x] \asymp \x^*,\qquad \text{if}\quad \tilde\gamma = \gamma
  \cdot \Big(1 - \frac{d_\gamma}s\Big),
\end{align*}
where $d_\gamma =\tr(\A^\top\A(\A^\top\A+\gamma\I)^{-1})$ is the
$\gamma$-effective dimension of $\A$ and $s$ is the sketch
size. Similar implicit regularization results have been shown 
for sketching-based  distributed regression \cite{dobriban2020wonder},
sampling-based ensemble methods \cite{lejeune2020implicit}, stochastic
optimization \cite{newton-less}, LR approximation \cite{precise-expressions,lejeune2024asymptotics} and randomized rounding \cite{dexter2024stochastic}.

\section{Putting Randomness into Next-generation Software}

\vspace{1mm}\noindent\textbf{Beyond BLAS/LAPACK to RandBLAS/RandLAPACK.}
One of the promises of RandNLA has been that it will lead to improved implementations in practice.
Indeed, the conclusion of the Blendenpik paper \cite{blendenpik}, in 2010, was that
``Blendenpik beats LAPACK's direct dense least-squares solver by a large margin on essentially any dense tall matrix,'' 
and that
``these results show the potential of random-sampling algorithms and suggest that random-projection algorithms should be incorporated into future versions of LAPACK.''%
\footnote{%
For readers not familiar with it, 
LAPACK (``Linear Algebra PACKage'') is a standard software library for NLA 
which provides routines for solving systems of linear equations and linear least squares, eigenvalue problems, and singular value decomposition, as well as routines to implement the associated matrix factorizations such as LU, QR, Cholesky and Schur decomposition.
It relies on an underlying BLAS (``Basic Linear Algebra Subprograms'') implementation to provide efficient and portable computational building blocks for its routines, where BLAS is a specification that prescribes a set of low-level routines for performing common linear algebra operations such as vector addition, scalar multiplication, dot products, linear combinations, and matrix multiplication \cite{templates}.
Importantly, although the BLAS specification is general, BLAS implementations are often optimized for speed on a particular machine, and they can bring substantial performance benefits, e.g., since BLAS implementations can take advantage of special floating point hardware.
}
This, in 2024, is timely: 
recent years have seen the convergence of several unprecedented changes, including formidable new system design constraints and revolutionary levels of hardware heterogeneity; and, 
due to these changes, much of the essential software infrastructure of computational science and engineering is, or will soon be, obsolete. 
These challenges motivated the BALLISTIC (``Basic ALgebra LIbraries for Sustainable Technology with Interdisciplinary Collaboration'') project \cite{ballistic_lawn}, which aims to enhance and update BLAS, LAPACK, and ScaLAPACK (``Scalable Linear Algebra PACKage'') by ``incorporating them into a layered package of software components ... that provides users seamless access to state-of-the-art solver implementations through familiar and improved Sca/LAPACK interfaces.''

As part of the BALLISTIC project \cite{ballistic_lawn}, 
we are introducing the use of RandNLA methods into BLAS and LAPACK, leading to RandBLAS and RandLAPACK \cite{randlapack_book_v2_arxiv}.
RandBLAS will be a library that concerns basic sketching for dense data matrices (with a reference implementation in C++).
Analogous to the original BLAS, it will be a ``contract'' between algorithm developers and hardware developers.
RandLAPACK will be a library that concerns algorithms for solving traditional linear algebra problems and advanced sketching functionality (to be written in C++ and built on BLAS++/LAPACK++ portability layer).
Analogous to the original LAPACK, it will consist of three main ``drivers'' (LS and optimization, LR approximation, and full-rank decompositions, in the case of RandLAPACK), and 
it will have ``computational routines'' (such as advanced sketching and error estimation).
The hope is that RandBLAS grows to become a ``community standard'' for RandNLA, more broadly than RandLAPACK, in the sense that its API would see wider adoption than any single implementation.

The connection with and value of Modern RandNLA for RandBLAS/RandLAPACK (and other such software projects) is subtle but important: 
\emph{due to the use of RMT-based ideas, in particular Def.~\ref{def:nearly-unbiased-estimator} of a nearly-unbiased estimator, we can obtain a much smaller theory-practice gap than was possible with Classical RandNLA, which depended on the notion of a subspace embedding from Def.~\ref{def:subspace_embedding}, for a broad range of implementations} \cite{debiasing-second-order,lacotte2020optimal,lacotte2022adaptive}.
For implementations in the past, one often used expensive Gaussian random projections to obtain stronger theory; and then implementations may or may not have used Gaussian random projections; thus leading to a potentially-large theory practice gap.
However, with the methods of Modern RandNLA, which with \eqref{prop:sub-gaussian} provide stronger control on the inversion bias, one can hope to analyze the distribution being implemented.

\vspace{1mm}\noindent\textbf{Beyond Single-machine Shared-memory.}
Going beyond the single-machine shared-memory setup, recent work within ML has focused on model training and inference: often with GPUs rather than CPUs; often in parallel/distributed setups; and often using low-precision computation.
Classical RandNLA methods have proven successful here, in particular in comparison to heavily hyperparameterized SGD methods \cite{yao2021adahessian}.
An emerging challenge with training and serving these models more generally involves memory and communication bottlenecks. 
In particular, ML applications are often bottlenecked by intra/inter-chip and communication across/to ML accelerators, rather than compute, a phenomenon referred to as the memory wall problem \cite{GYKx24_memory_wall_TR}.
Quantization is one solution \cite{quantiz_review_TR21}, and recent work (using Modern RandNLA theory) suggests the promise of using RandNLA methods here too \cite{ZCM20_quantized_TR}.

\section{Concluding thoughts}

There are many topics we did not cover:
data-driven methods for learning good sketching operators \cite{indyk2019learning,indyk2021few,cho2023surrogate};
RandNLA methods for tensor decompositions \cite{wang2015fast,li2017near,erichson2020randomized,jin2021faster};
RandNLA methods in streaming/online environments \cite{braverman2020near};
probabilistic numerics \cite{bartels2019probabilistic}; 
as well as topics such as
Hutchinson and spectral function methods \cite{fred_implicit_trace_15,yao2021adahessian},
using randomness deep inside an algorithm, e.g., for pivot rule decisions,  
using these methods for the theory/practice of neural networks, e.g., NTK and Nystromformer,  
using these methods for quantization and low-precision computation \cite{quantiz_review_TR21,GYKx24_memory_wall_TR,dexter2024stochastic},
and using these methods to deal with graph-related and other structured matrices such as Laplacians \cite{DM10_TR,KMP12,Mah16_SGM_TR}.
In many of these cases, it is increasingly-important to identify core linear algebraic structures and build algorithmic/statistical methods around them (rather than ``tacking'' them on later as a ``band aid'' to fix problems that arise when one ignores key linear algebraic issues).
This will be increasingly-important as heterogeneous hardware trends and ML training trends continue to grow.
We expect the ``Modern'' theoretical tools developed by/for RandNLA that we have covered will be useful in these and other~situations.

\subsubsection*{Acknowledgments}
MD would like to acknowledge NSF CAREER for partial support.  MWM would like to
acknowledge the NSF and ONR via its BRC on RandNLA for partial
support of this work.

\bibliographystyle{plain}
\bibliography{../../pap.bib}

\end{document}